\documentclass{article}

\usepackage{booktabs}
\usepackage{threeparttable}
\usepackage{colortbl}
\DeclareUnicodeCharacter{2212}{-}

\usepackage{bm,url}
\usepackage{times}
\usepackage{epsfig,subfigure}
\usepackage{graphicx}
\usepackage{amsmath,amsthm}

\usepackage{amssymb}
\usepackage{enumerate}
\usepackage[algoruled,vlined]{algorithm2e}
\usepackage{color}
\usepackage{tabu,multicol}
\usepackage{makecell,tikz}
\usepackage[switch,columnwise]{lineno}

\usepackage{longtable}
\usepackage{chngpage}
\usepackage{booktabs}
\usepackage{hhline}

\renewcommand{\raggedright}{\leftskip=0pt \rightskip=0pt plus 0cm}

\usepackage{multirow}

\DeclareMathOperator*{\argmin}{argmin}
\newcommand{\A}{\mathbf{A}}
\newcommand{\B}{\mathbf{B}}

\newcommand{\I}{\mathbf{I}}

\newcommand{\U}{\mathbf{V}}

\newcommand{\x}{\mathbf{x}}

\newcommand{\Thetab}{\boldsymbol{\Theta}}
\newcommand{\thetab}{\boldsymbol{\theta}}

\usepackage{tabularx}

\title{A Survey of Geometric Optimization for Deep Learning: From  Euclidean Space to Riemannian Manifold}

\author{%
  Yanhong Fei, Xian Wei, Yingjie Liu, Zhengyu Li, Mingsong Chen\\
  Software Engineering Institute, East China Normal University
 } 

\begin{document}

\maketitle

\begin{abstract}
Although Deep Learning 
(DL) has achieved success in
complex Artificial Intelligence (AI) tasks, 
it suffers from various notorious problems (e.g., 
feature redundancy, and vanishing or exploding gradients),
since updating parameters in Euclidean space cannot fully exploit the geometric structure of the solution space. 
%
As a promising alternative solution, 
Riemannian-based DL uses geometric optimization to update parameters on Riemannian manifolds and can leverage the underlying geometric information. 
Accordingly, this article presents a comprehensive survey of 
applying geometric optimization in DL. 
%
%
At first, this article introduces the basic procedure of 
the geometric optimization, 
including various geometric optimizers and some concepts of Riemannian manifold. 
Subsequently, this article investigates the application of 
geometric optimization in different DL networks in various AI tasks, 
e.g., convolution neural network, recurrent neural network, transfer learning, and optimal transport.
 %
%
Additionally, typical public toolboxes that implement optimization on manifold are also discussed.  
%
Finally, this article makes a performance comparison between different deep geometric optimization methods under image recognition scenarios.
%

\end{abstract}




\maketitle

\section{Introduction}



With increasing computing power, deep neural networks optimized in Euclidean space have achieved remarkable success 
from computer vision to natural language processing (e.g., autonomous driving and protein structure prediction)
 \cite{Goodfellow_Bengio2016_Book,townshend2021geometric_Science}. 
However, to fully exploit the valuable information hidden in the data, most deep learning models tend to increase the capacity of their networks,  either by  widening the existing layers or by adding more layers \cite{tan2019efficientnet_icml, he2017wider_nips, rolnick2018power_iclr}. 
For example,  models  often contain hundreds of convolution and pooling layers with various activation functions and multiple fully connected layers, 
producing millions or billions of parameters 
during training.
These massive parameters associated with complex model architectures 
challenge the optimization of deep learning networks.
As an alternative paradigm, optimization on the Riemannian manifold exploits hidden valuable information by utilizing geometric properties of parameters, rather than increasing the network capacity. 
Therefore, geometric optimization can alleviate over-parameterization and feature redundancy problems.
For example, deep learning models trained on the orthogonal
manifold have less correlated parameters, making features much less redundant \cite{Wang2020OrthogonalCN}. 

The optimization objective in most deep learning methods can be formulated as 
\begin{equation}
\label{eq:1}
\begin{aligned}
&  \argmin_{\thetab \in \mathcal{D}} \ f_{\thetab} (\x),\  s.t.\ C (\thetab),
\end{aligned}
\end{equation}
where $\mathcal{D}$ denotes the predefined admissible search space,
$f$ denotes a real-value optimization function (e.g., loss function) to be minimized by trainable parameters $\thetab$, and $C (\thetab)$ represents  constraints (e.g., orthogonality \cite{Wang2020OrthogonalCN} and unit row sums  \cite{2021Variational}) that $\thetab$ is subject to.
Most deep learning methods define the search space $\mathcal{D}$ as the Euclidean space.
%
However, parameters satisfying constraints are on the manifold, which is a low dimensional subspace and only occupies a small part of Euclidean space.
Therefore, to eliminate constraints and reduce parameters, 
geometric optimization \cite{absi:book08,boumal2020intromanifolds_opti,hu2020brief,Wei2017Thesis} narrows the search space from Euclidean space to a smooth manifold. 
%
Hence, 
Equation~\eqref{eq:1} is transformed into a differentiable optimization function $f: \mathcal{M} \rightarrow \mathcal{R}$ on a Riemannian manifold, i.e., 
\begin{equation}
\label{manifold_optimization}
    \argmin_{\theta \in \mathcal{M}, M = \{\theta | C(\theta)\}} \ f_{\thetab} (\x).
\end{equation}
As shown in Equation~\eqref{manifold_optimization}, selecting a manifold composed of points that meet constraints $C (\thetab)$ in Equation~\eqref{eq:1}, a large class of constrained deep learning problems in Euclidean space can be optimized as unconstrained and convex ones on the Riemannian manifold \cite{hu2020brief}, which helps ensure the convergence.
For example, a typical dimension reduction problem can be defined as follows
\begin{equation}
\label{constrained_DR}
\begin{aligned}
& \argmin_{\thetab \in {E}} f_{\thetab} (\x) = − tr (\thetab^{T} x^{T} x\thetab),\ s.t.\ \ \thetab^T\thetab=I,
\end{aligned}
\end{equation}
where $E$ represents the Euclidean space, $I$ represents the identity matrix and  parameters $\thetab$ are constrained to be orthogonal. 
Since all matrices that satisfy orthogonality compose of the \textit{Stiefel} manifold , Equation~\eqref{constrained_DR} can be treated as an unconstrained problem on the \textit{Stiefel} manifold, which is a kind of Riemannian manifold.
%

%
%
\begin{figure}[h]
\centering
\includegraphics[width=2.5in]
{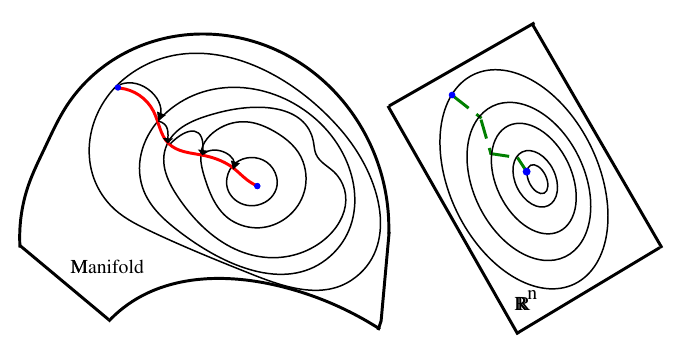}
\caption{
\label{fig:fig1}
Comparison between geometric and
Euclidean optimization path. The blue center point is the global optimum. The red curve describes the Riemannian optimization path converging upon the global optimal goal, always along a curve on manifolds. In contrast, the green dotted line indicates the Euclidean gradient descent path towards the optimal goal, taking the risk of moving off the manifold. 
}
\end{figure}
Figure~\ref{fig:fig1} depicts the intuitive paradigm for optimization processes in arbitrary Euclidean space and on Riemannian manifolds.
Traditional optimization methods in Euclidean space may ignore the advantages of applying geometric optimization strategies.
For example,
the latter can obtain richer geometric information from different unique manifold structures and convert constrained optimization problems into unconstrained problems.
Moreover, geometric optimization can achieve faster convergence speed and mitigate gradient explosion and disappearance problems in deep learning, which will be detailed in Section~\ref{sec:04}.
Due to the above potential, geometric optimization has been applied to various deep neural networks in recent years,
such as convolution neural network  (CNN)  \cite{Bansal2018CanWG,Wang2020OrthogonalCN,lezcano2019cheap}, recurrent neural network  (RNN)  \cite{2015Unitary} and vision transformer (ViT) \cite{fei2022vit}.
For instance, orthogonal parameterization is used in CNN to reduce filter similarities, make spectra uniform \cite{2008IEEE}, and stabilize the activation distribution in different network layers \cite{P2016Regularizing}.
However, there is a lack of comprehensive surveys focused on deep learning methods applying geometric optimization.
To explore benefits of geometric optimization, this article aims to give an overall review of recent advances on applying geometric optimization in deep learning. 

\begin{figure}[h]
\centering
\includegraphics[width=5.5in]
{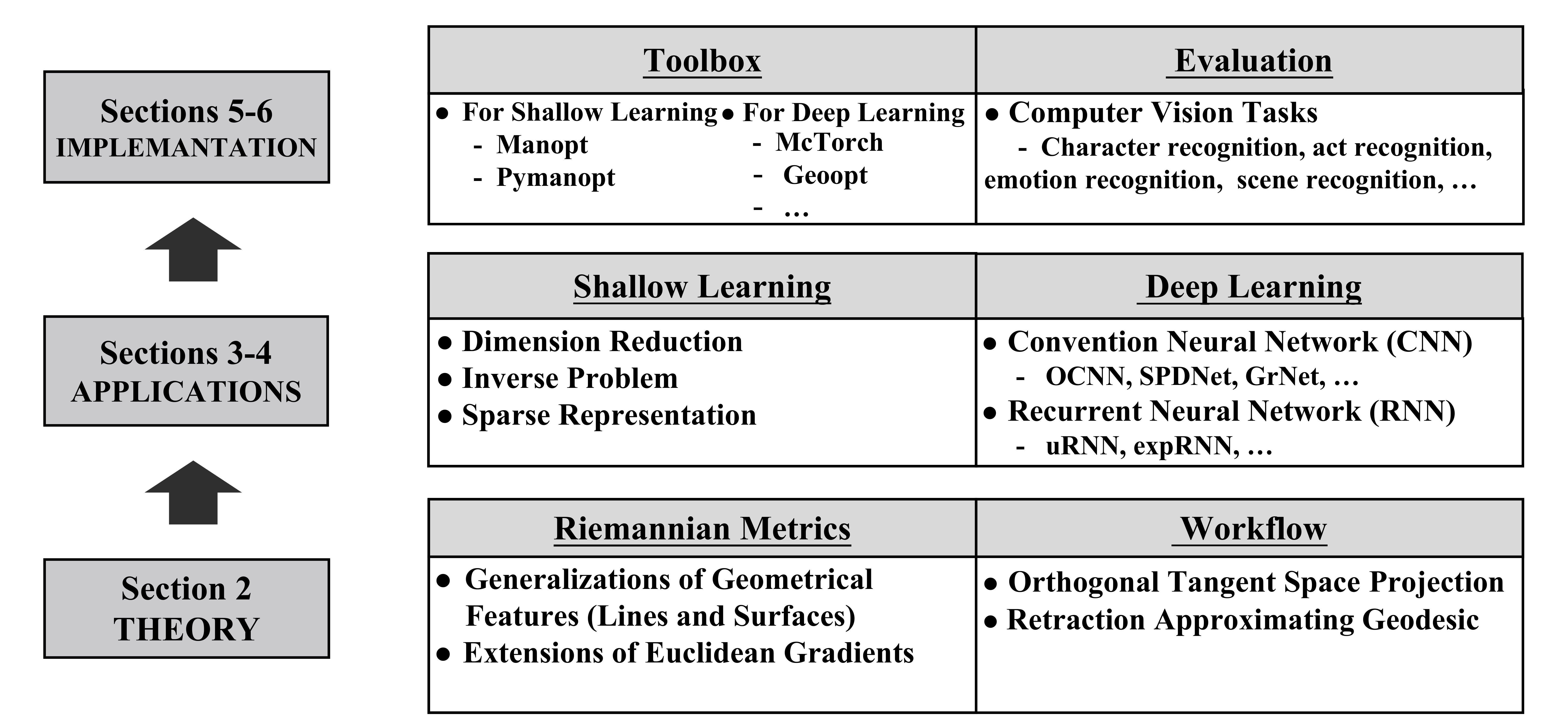}
\caption{
\label{bird_eye_view}
An overview of the central idea of this article.
}
\end{figure}
\footnote{\textbf{\emph{Notations}} In this work, vectors and matrices are denoted by bold lower case letters and upper case ones, respectively. 
Let 
$\mathbb{R}$ be the set of real numbers, $\mathbb{C}$ be the set of complex numbers and
$\nabla f$ denotes the Euclidean gradient.}
\textbf{Overview and article organization. }
In this article, a survey of geometric optimization techniques for deep learning is presented, including the theory and applications of geometric optimization.
Figure~\ref{bird_eye_view} displays an overview of the central idea of this article.
Since the optimization theory is unified and model-independent, this article illustrates the theory first, including various geometric gradient descent optimizers (Section~\ref{sec:02}).
The motivation and technique of applying geometric optimization in classical machine learning is different from that of deep learning. Therefore, this article reviews how to apply geometric optimization to shallow learning (Section~\ref{sec:03}) and deep learning (Section~\ref{sec:04}) separately. 
In particular, this article investigates representative manifold optimization toolboxes (Section~\ref{sec:05}), followed by performance comparisons of different geometric deep learning methods on image recognition tasks (Section~\ref{sec:06}).  
Finally, we conclude the article and highlight future challenges and research trend (Section~\ref{sec:conclu}). 

%

\section{Geometric Optimization Theory}
\label{sec:02}
The essence of an optimization problem is to find the maximum or minimum value of a cost function. 
An unconstrained optimization problem 
can use conventional optimization methods (e.g., steepest descent method, 
conjugate gradient method, and Newton method) to find an optimal solution \cite{nocedal2006numerical_book}.
However, a broad range of optimization problems that occur in computer vision tasks are known as constrained optimization problems.
In such a case, finding a closed form for the cost function is difficult.
To use the aforementioned conventional optimization techniques, 
the constrained problem can be transformed into an unconstrained form by
using the method of Lagrange multipliers or using a barrier penalty function  \cite{nocedal2006numerical_book}.
%
However, the above methods hardly take advantage of underlying manifold structures. 
They merely treat the constrained problem as a ``black box'' and solve them by using algebraic manipulation.

As an alternative solution, 
geometric optimization methods
are developed to exploit intrinsic geometric structures of objective function parameters.
By utilizing the underlying geometry of a cost function, geometric optimization methods can narrow the search space of constrained optimization problems from Euclidean space to smooth Riemannian manifolds. 
Riemannian manifold has a differentiable structure and is equipped with smooth inner product and Riemannian gradients, which are different from Euclidean space and lay the foundation for geometric optimization.
Based on the Riemannian inner product and Riemannian gradients, a broad spectrum of conventional optimization techniques in Euclidean space can have their counterparts on smooth manifolds \cite{luenberger1984linear_book, gabay1982minimizing_jota, brockett1993differential_spma, absi:book08}, 
including the steepest descent method  \cite{luenberger1984linear_book, gabay1982minimizing_jota, brockett1993differential_spma}, 
conjugate gradient descent method  \cite{edelman1998geometry_siam, hawe2013separable_cvpr,wei2016trace_cvpr}, trust-region method  \cite{absi:book08, absil2007trust_fcm} 
and Newton’s method  \cite{dedieu2003newton_jna, absi:book08}.
Therefore, geometric optimization methods can use Riemannian optimizers to find an optimal solution for objective functions.

In the following subsections, this article first illustrates the model-independent  optimization process on the Riemannian manifold, covering basic concepts related to geometric optimization  (Section~\ref{2.1}).
Next, this article briefly introduces various Riemannian gradient descent optimizers implementing geometric optimization, which is a counterpart of optimizers in Euclidean space (Section~\ref{2.2}).
Finally, this article presents a series of manifold structures that are commonly used in deep geometric learning methods (Section~\ref{sec:intro_manifolds}).
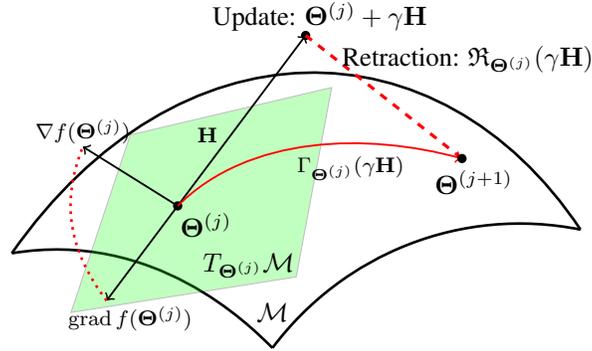
\begin{figure}[h!]
\begin{center}
	\begin{tikzpicture}[scale=0.9, inner sep=0pt,dot/.style={fill=black,circle,
		minimum size=7pt},scale=0.35]
	\draw[line width=1pt]  (-9,0) .. controls  (-2,10) and  (9,10) ..  (15,1);
	\draw[line width=1pt]  (-9,0) .. controls  (-3,1) and  (1,-3) ..  (2,-4);
	\draw[line width=1pt]  (2,-4) .. controls  (5,-0) and  (10,2) ..  (15,1);
	\node[] at  (2.0,-2.5) {$\mathcal{M}$};
	\draw[fill=green, opacity=0.25]  (-6.5,-2.5) --  (-4,5) --  (4.5,7) --  (3,-1) -- cycle;
	\node[] at  (-0.8,1.2) {$\bm{\Theta}^{ (j)}$};
	\node[] at  (1,-0.5) {$T_{\bm{\Theta}^{ (j)}} \mathcal{M}$};
	\node[dot,scale=0.5]  (xk) at  (-2,2) {};
	%
        \draw[-{>[scale=2.5]},line width=0.7pt]  (-2,2) --  (3.4, 9.2);
	\node[dot,scale=0.5]  (xk1) at  (3.4, 9.2) {};
	\node[] at  (-0.75, 5) {\footnotesize $\mathbf{H}$}; 
	\node[] at  (4.0, 10.0) {Update: $\bm{\Theta}^{ (j)} + \gamma \mathbf{H}$};
	\draw[red, dashed, very thick]  (3.4, 9.2) --  (10,4);  
	\node[] at  (10.25, 8.2) {Retraction: $\mathfrak{R}_{\bm{\Theta}^{ (j)}} (\gamma \mathbf{H})$};
 \draw[-{>[scale=2.5]},line width=0.7pt]  (-2,2) --  (-5,-2);
	\node[] at  (-4,-2.75) { \footnotesize $\operatorname{grad}{f} (\bm{\Theta}^{ (j)})$ };
	\draw[-{>[scale=2.5]},line width=0.7pt]  (-2,2) --  (-6.0, 4.5);
	\node[] at  (-6,5.15) {\footnotesize $\nabla{f} (\bm{\Theta}^{ (j)})$ };
	\draw[red, dotted, line width=1pt]  (-5,-2) .. controls  (-6.5,0) and  (-7,2) ..  (-6.0, 4.5);
	%
 \draw[-{>[scale=3.5, red]},red,line width=0.7pt]  (-2,2) .. controls  (1,5) and  (6,5.1) ..  (10,4);
	%
	\node[] at  (5.3,3.6) {\footnotesize 
		$\Gamma_{\bm{\Theta}^{ (j)}} (\gamma \mathbf{H})$ 
	};  
	\node[] at  (10.5,3.0) {$\bm{\Theta}^{ (j+1)}$};
	\node[dot,scale=0.5]  (xk1) at  (10,4) {};
	\end{tikzpicture}
\end{center}
\caption{The update process in geometric gradient descent algorithm. 
	It shows an update from the point $\bm{\Theta}^{ (j)}$ to the point  $\bm{\Theta}^{ (j+1)}$ in a search direction 
	$\mathbf{H} \in T_{\bm{\Theta}^{ (j)}} \mathcal{M}$ along the geodesic curve $\Gamma_{\bm{\Theta}^{ (j)}} (\gamma \mathbf{H})$.
	%
	Moreover, it   describes how to approximate the geodesic $\Gamma_{\bm{\Theta}^{ (j)}} (\gamma \mathbf{H})$ by using the retraction $\mathfrak{R}_{\bm{\Theta}^{ (j)}} (\gamma \mathbf{H})$.
}
\label{fig_3_GD}
\end{figure}

\subsection{Geometric Optimization Process on Manifolds}
\label{2.1}
Figure~\ref{fig_3_GD} depicts the update process in  geometric optimization\cite{Wei2017Dynamic_TIP} through the gradient descent example.
%
There are two nearby points $\Theta^{ (j)}$ and $\Theta^{ (j+1)}$ on a manifold $\mathcal{M}$
together with the tangent space at $\Theta^{ (j)}$ (refer to the green area in Figure~\ref{fig_3_GD}). Each point $\Theta$ on the manifold has its corresponding tangent space $T_{\Theta} \mathcal{M}$,
which is a generalization of the tangent plane in Euclidean space and consists of all tangent vectors passing through  $\Theta$  \cite{hawe2013learning}. 
%
Each tangent space has an inner product, which is vital for vector metrics such as length and angles. 
%
Inner product space further helps induce the concept of orthogonality, an extension of vertical in higher dimensions. 
A Riemannian gradient $grad\, f(\Theta)$ for geometric optimization is a tangent vector on the tangent space $T_{\Theta} \mathcal{M}$ and points to the direction where the cost function on the manifold ascends steepest  \cite{hawe2013learning}. 
Figure~\ref{fig_3_GD} shows that gradient $\nabla f(\Theta)$ is computed in the ambient Euclidean space.
Since the manifold is locally homomorphic to the Euclidean space, 
$grad\, f(\Theta)$ can be achieved by projecting Euclidean gradient $\nabla f(\Theta)$ to the appropriate tangent space $T_\Theta \mathcal{M}$, i.e., 
\begin{equation}
    grad\ f(\Theta) = \Pi _{T_\Theta \mathcal{M}} (\nabla f(\Theta)),
\end{equation}
where $\Pi$ means the orthogonal projection.

As a counterpart of Euclidean straight lines, 
a geodesic is a locally shortest path between two points on the manifold. Therefore, reaching the optimal goal along a correct geodesic is shortest. 
Formally, a geodesic $\Gamma_{\bm{\Theta}} (\gamma \mathbf{H})$ is a smooth curve on the manifold, proceeding from $\Theta$ in the direction of tangent vector $\mathbf{H} \in T_\Theta \textit{M}$ with a step size of $\gamma \in \mathbb{R^+}$ \cite{hawe2013separable_cvpr}. 
Since each tangent vector is the direction vector of a specific geodesic curve, it can uniquely determine a geodesic curve. 
In particular, the geodesic defined by the negative Riemannian gradient reveals the next point in the optimization direction.
A point can be mapped from the tangent space to the manifold through exponential mapping. 
In practice, to alleviate the high computational cost of exponential mapping, retraction operation  $\mathfrak{R}_{\bm{\Theta}} (\gamma \mathbf{H})$
is often used as an approximation  \cite{2018Geometry}:
\begin{equation}
    \mathfrak{R}_{\bm{\Theta}} (\gamma \mathbf{H}): T_{p}\mathcal{M} \rightarrow \mathcal{M},\ \gamma \mathbf{H} \rightarrow \Gamma_{\bm{\Theta}} (\gamma \mathbf{H}),
\end{equation}
where $\mathbf{H}$ denotes an opposite vector of the Riemannian gradient. 
Therefore, $\mathbf{H}$ points in the direction of the steepest descent of the optimization function.
As a result, the optimization function will be minimized if parameter $\Theta$ is updated along a geodesic curve in the direction of $\mathbf{H}$. In summary, with a step size of $\gamma$, the optimizing process from the current parameter $\Theta^{ (j)}$ to the next parameter $\Theta^{ (j+1)}$ can be formulated as
\begin{equation}
     \Theta^{ (j+1)} = \Gamma_{\bm{\Theta}^{(j)}} (\gamma \mathbf{H}) \approx \mathfrak{R}_{\bm{\Theta}^{(j)}} (\gamma \mathbf{H}) = \mathfrak{R}_{\bm{\Theta}^{(j)}} (-\gamma grad\, f(\Theta^{(j)})).
\end{equation}


\subsection{Gradient Descent Optimizers}
\label{2.2}
Optimization problems defined in Euclidean space can be abstracted as
\begin{equation}
	\label{decent_1}
	{\min} \{f_{\thetab} (\x):\thetab \in \mathbb{E}\},
\end{equation}
where $\thetab$ are trainable parameters and $E$ means the Euclidean space.
There are a variety of standard optimizers for Equation~\eqref{decent_1}.
The gradient descent method is a most basic optimization strategy. It can be improved by stochastic gradient descent (SGD), which can accelerate convergence. 
The other two typical variants of the gradient descent method are stochastic gradient descent-momentum (SGD-M) and root mean square prop (RMSProp).
To solve valley oscillation and saddle point stagnation problems that SGD suffers from, SGD-M is developed to maintain the inertia of the previous step. 
According to empirical judgments of different parameters, RMSProp can adaptively determine the learning rate of parameters, i.e., parameters with low update frequency can have a larger learning rate, while parameters with high update frequency can reduce the step size.
Let $\thetab^{(k)}$ represent parameters at iteration $k$ and $\thetab^{(k+1)}$ represent parameters at iteration $k+1$, this section first explains the above Euclidean gradient descent optimizers and then shows how to generalize them to the Riemannian manifold for geometric optimization \footnote{For simplicity, this paper uses ${\nabla}f$ to denote $\frac{\partial f_{\thetab}(x)}{\partial \thetab}$ in Section~\ref{2.2}.}. 

\textbf{Gradient Descent}.	
The gradient descent method takes the following form
	\begin{equation}
	\label{decent_2}
	\thetab^{(k+1)} = \thetab^{(k)} - {\lambda}{\nabla}f(\thetab^{(k)}),
	\end{equation}
where $\lambda$ is a hyper-parameter representing the step size.
The negative direction of the gradient ${\nabla}f(\thetab^{(k)})$ has a vital property, i.e., it is a descent direction of the optimization problem. Therefore, the optimization process is to iteratively update trainable parameters along the negative direction of gradient until convergence.

\textbf{Stochastic Gradient Descent (SGD)}.
The main idea behind SGD is to use random mini-batches of training data to update parameters of the optimization problem, which inherently reduces the calculation workload.
Although the parameters may not be updated in the direction of the steepest descent every time, the overall update is in the steepest descent direction through multiple rounds of updates.
As a result, SGD can greatly speed up the optimization process.

\textbf{Stochastic Gradient Descent-Momentum (SGD-M)}.
Inspired by the concept of momentum in physics, SGD-M exerts the influence of the last update on the current update to damp oscillation and accelerate convergence.
Let $m^{ (k)}$ denote the update imposed on $\thetab^{ (k-1)}$ and $\nabla f$ denote the gradient at time $k$, the update $m^{ (k+1)}$ to be imposed on $\thetab^{ (k)}$ can be achieved as
\begin{equation}
\label{sgd-m-1}
    m^{ (k+1)} = \lambda_0\ m^{ (k)} + \lambda_1 \nabla f,
\end{equation}
where $\lambda_0$ and $\lambda_1$ are hyper-parameters.
Sequentially, the parameter $\thetab^{ (k)}$ is updated to $\thetab^{ (k+1)}$ by $m^{ (k+1)}$ as follows
\begin{equation}
\label{sgd-m-2}
    \thetab^{ (k+1)} = \thetab^{ (k)} - m^{ (k+1)}.
\end{equation}

\textbf{Root Mean Square Prop (RMSProp)}.
Similar to SGD-M, RMSProp considers the influence of the last update when calculating the upcoming update. Let $m^{ (k)}$ be the update on the previous occasion and $\nabla f$ be the current gradient, RMSProp designs upcoming update $m^{ (k+1)}$ as follows
\begin{equation}
\label{rms-1}
    m^{ (k+1)} = \lambda\ m^{ (k)} +  (1-\lambda) (\nabla f \odot \nabla f),
\end{equation}
where $\lambda$ is a hyper-parameter and $\odot$ denotes the $Hadamard$ product  \cite{2018Geometry} which is element-wise. RMSProp updates $\thetab^{ (k)}$ to $\thetab^{ (k+1)}$ in the following way, i.e.,
\begin{equation}
\label{rms-2}
    \thetab^{ (k+1)} = \thetab^{ (k)} - \eta \frac{\nabla f}{\sqrt{m^{ (k+1)}+\epsilon}},
\end{equation}
where $\eta$ is a hyper-parameter and $\epsilon$ is positive to prevent the denominator from being zero.
Using element-wise square root and division operation, RMSProp guarantees that different elements in gradient $\nabla f$ have different coefficients, which represent learning rates in deep learning. Therefore, RMSProp enables parameters to have different learning rates  \cite{2018Geometry}, which makes the optimization process more flexible.

Based on the aforementioned optimization process on manifolds (Section~\ref{2.1}), the Euclidean gradient descent algorithm in Equation~\eqref{decent_2} can be transferred to Riemannian manifolds as 
\begin{equation}
    \theta^{ (k+1)} =  \mathfrak{R}_{{\theta}^{(k)}} (-\lambda grad\, f(\theta^{(k)})),
\end{equation}
where $\mathfrak{R}_{{\theta}^{(k)}}$ means the retraction operation at point ${\theta}^{(k)}$ and $grad\, f$ means the Riemannian gradient.
For better understanding, this article takes  constraint SGD-M and constraint RMSProp as an instance to explain how to generalize gradient descent optimizers from Euclidean space to manifolds. 
By performing orthogonal projection and retraction, other Euclidean gradient descent optimizers can be similarly converted to Riemannian optimizers.

 \textbf{Constraint SGD-M} \cite{2018Geometry}.  
Constraint SGD-M is a generalization of SGD-M optimizer on manifolds.
In the $k$-th iteration, $m^{ (k)}$ denotes a tangent vector on the tangent space $T_{\thetab^{ (k-1)}}M$ and $m^{ (k+1)}$ denotes another vector on the tangent space $T_{\thetab^{ (k)}}M$. Since $\nabla f$ is in the surrounding Euclidean space, it needs to be orthogonally projected to tangent space $T_{\thetab^{ (k)}}M$, 
i.e., the current Riemannian gradient $grad\, f$ is achieved as follows
    \begin{equation}
        grad\, f = \Pi_{T_{\thetab^{ (k)}}M} (\nabla f).
    \end{equation}
The transportation from a tangent space associated with point $p$ to another tangent space associated with point $q$ is called parallel transportation, i.e., $\Gamma_{p\rightarrow q}:\ T_pM \rightarrow T_qM$.
After projecting the Euclidean gradient $\nabla f$ to the tangent space $T_{\thetab^{ (k)}}M$ and transporting $m^{ (k)}$ from $T_{\thetab^{ (k-1)}}M$ to $T_{\thetab^{ (k)}}M$, Equation~\eqref{sgd-m-1} is transformed to: 
    \begin{equation}
        m^{ (k+1)} = \lambda_0 \, \Gamma_{\thetab^{ (k-1)}\rightarrow \thetab^{ (k)}} (m^{ (k)}) + \lambda_1 \, \Pi_{T_{\thetab^{ (k)}}M} (\nabla f).
    \end{equation}
Based on the retraction operation, the optimization parameter $\thetab^{ (k+1)}$ can be updated from $\thetab^{ (k)}$ by searching along the geodesic in the negative direction of $m^{ (k+1)}$, i.e., the iterate optimization can be expressed as
    \begin{equation}
        \thetab^{ (k+1)} = \mathfrak{R}_{\thetab^{ (k)}} (-m^{ (k+1)}).
    \end{equation}
    
\textbf{Constraint RMSProp}  \cite{2018Geometry}.
Similar to constraint SGD-M, after transporting $m^{ (k)}$ from tangent space $T_{\thetab^{ (k-1)}}M$ to $T_{\thetab^{ (k)}}M$ and orthogonally projecting $\nabla f \odot \nabla f$ to corresponding tangent space, Equation~\eqref{rms-1} can be transformed into:
    \begin{equation}
        m^{ (k+1)} = \lambda \Gamma_{\thetab^{ (k-1)}\rightarrow \thetab^{ (k)}} (m^{ (k)})
        + (1-\lambda)\Pi_{T_{\thetab^{ (k)}}M} (\nabla f \odot \nabla f).
    \end{equation}
The parameter $\thetab^{ (k+1)}$ of the optimization goal can be iteratively searched on the manifold with a determined direction $-\eta\,\frac{\Pi_{T_{\thetab^{ (k)}}M} (\nabla f)}{\sqrt{m^{ (k+1)}+\epsilon}}$, that is,
    \begin{equation}
        \thetab^{ (k+1)} = \mathfrak{R}_{\thetab^{ (k)}} (-\eta\,\frac{\Pi_{T_{\thetab^{ (k)}}M} (\nabla f)}{\sqrt{m^{ (k+1)}+\epsilon}}).
    \end{equation}

\subsection{Manifold Examples}
\label{sec:intro_manifolds}

Different kinds of matrix manifolds have different geometry structures and satisfy different constraints, bringing different advantages when applying geometric optimization to deep learning.
For example, the oblique manifold plays a significant role in dictionary learning due to its property of unit-norm columns, while the \textit{Stiefel} manifold has a positive effect on optimizing RNNs since matrices on  the \textit{Stiefel} manifold have orthogonal and uncorrelated columns, which helps alleviate feature abundancy problems in RNNs.
Since space is limited, this section only presents common manifold structures\footnote{For more introduction on matrix manifolds, we refer interested readers to the website https://www.Pymanopt.org.} such as \textit{Stiefel} manifold, oblique manifold, and \textit{Gra{\ss}mann} manifold, all of which are widely used in existing geometric optimization techniques that are discussed in Section~\ref{sec:03} and Section~\ref{sec:04}.


\textbf{Product Manifold and Quotient Manifold.}	
Let $\mathcal{A}$ and $\mathcal{B}$ be two manifolds of dimension $d_A$ and $d_B$, for any pair of charts $ (U,\phi)$ and $ (V,\varphi)$ of $\mathcal{A}$ and $\mathcal{B}$, the map $\Phi$ is defined on $U\times V$ by $\Phi (x,y) =  (\phi (x),\varphi (y))$.
It specifies a smooth product manifold structure on the product space $\mathcal{A}\times \mathcal{B}$. 
Quotient manifold is an abstract space with similar subsets in the same manifold.
These subsets can be described with equivalence relationship.
$\mathcal{A}$ represents a manifold equipped with an equivalence relation $\sim$, 
which satisfies three properties, i.e., reflexivity, symmetry and transitivity \cite{absi:book08}.
The equivalence class of one point $x$ consists of all elements that are equivalent to it, i.e.,
\begin{equation}
    [x] := \{y \in \mathcal{A} : y \sim x \},
\end{equation}
where $[x]$ indicates the equivalence class of $x$.
The quotient of manifold $\mathcal{A}$ by relation $\sim$ is defined as follows 
    \begin{equation}
        \mathcal{A}/\sim \ := \{[x] : x\in \mathcal{A}\},
    \end{equation}
with the projection $\pi: \mathcal{A} \rightarrow \mathcal{A}/\sim$, indicated by $x \rightarrow [x]$.
When $\pi$ is a submersion projection, 
and $\mathcal{A}$ is a smooth manifold \cite{absi:book08,lee2013smooth},
$\mathcal{A}/\sim$ admits a unique smooth manifold structure $B$, which is the quotient manifold of $\mathcal{A}$.
%

\textbf{Symmetric Positive-Definite Manifold} \cite{2018Geometry}. 
It consists of Symmetric Positive-Definite  (SPD) matrices $M \in \mathbb{R}^{p \times p}$ equipped with the Affine Invariant Riemannian Metric  (\emph{AIRM}) as follows
    \begin{equation}
        S_{++}^p \triangleq \{M \in \mathbb{R}^{p \times p}: v^TMv > 0, \forall v\in \mathbb{R}^p - \{0_p\}\}.
    \end{equation}
SPD manifold achieves great success in computer vision due to its powerful statistical representations for images and videos. For example, SPD matrices are used to construct region covariance matrices for pedestrian detection \cite{Tosato}, 
    joint covariance descriptors for action recognition \cite{harandi2014manifold},
    and image set covariance matrices for face recognition \cite{huanga15}.

\textbf{\textit{Stiefel} Manifold}  \cite{2018Geometry}.
The \textit{Stiefel} manifold $St (p, n)$ is composed of orthogonal matrices $W \in \mathbb{R}^{n \times p}  (p \le n)$ endowed with the Frobenius inner product as follows
    \begin{equation}
        St (p, n) \triangleq \{W \in \mathbb{R}^{n \times p}: W^TW = I_p\},
    \end{equation}
where $I_p$ denotes $\mathbb{R}^{p \times p}$ identity matrix. The optimization function over the compact \textit{Stiefel} manifolds has an upper bound, which allows it to achieve an optimal solution.

\textbf{Sphere Manifold and Oblique Manifold.}   
The set of unit Frobenius norm matrices of size $n\times m$ is denoted by the sphere $\mathbb{S}^{nm-1}$.
It can be treated as a Riemannian submanifold embedded in Euclidean space $\mathbb{R}^{n\times m}$ endowed with the usual inner product $\langle H_1, H_2 \rangle = \operatorname{trace} (H_1^T H_2)$.
The oblique manifold $\mathcal{OB} (n, m)$ is the set of matrices of size $n\times m$ with unit-norm columns.
It has  the same geometry  as that of the product manifold of spheres $\prod_{i=0}^{m} \mathbb{S}^{n-1}$.
    
\textbf{\textit{Gra{\ss}mann} Manifold} \cite{2018Geometry}.  The \textit{Gra{\ss}mann} manifold $\mathcal{G} (n, p)$ embraces the set of subspaces spanned by the orthogonal matrices $X \in \mathbb{R}^{n \times p}  (p \le n)$ as
    \begin{equation}
        \mathcal{G} (n,p) \triangleq \{Span (X): X \in \mathbb{R}^{n \times p}, X^TX = I_p\}.
    \end{equation}
Note that a \textit{Gra{\ss}mann} manifold is different from a \textit{Stiefel} manifold, i.e., a point on the \textit{Stiefel} manifold represents a basis for a subspace, whereas a point on the \textit{Gra{\ss}mann} manifold represents an entire subspace. Moreover, \textit{Gra{\ss}mann} manifolds are of linear subspaces and can be used to perform a geometry-aware dimension reduction.
    
\textbf{{Unitary} Manifold.}    
Unitary matrices are the extension of orthogonal matrices to the complex domain, i.e.,
    \begin{equation}
        U (n) \triangleq \{U \in \mathbb{C}^{n \times n}: U^{\ast}U = I_n\},
    \end{equation}
where $U^{\ast}$ denotes the conjugate transpose matrix and $I_n$ represents the identity matrix of size $n\times n$.  Orthogonal or unitary matrices can preserve norm of vectors, i.e., $\Vert Wh \Vert_{2}$ = $\Vert h \Vert_{2}$ when $W$ is an orthogonal or unitary matrix.
Therefore, exploding and vanishing gradient problems in deep temporary networks can be alleviated when parameters are optimized on the orthogonal or unitary manifold, which will de detailed in Section~\ref{sec:4.2}.
    
\textbf{Lie Group}  \cite{lezcano2019cheap}.
Lie groups are real or complex manifolds with group structure. There are two compact and connected Lie groups, i.e., the special orthogonal group formulated as
    \begin{equation}
    SO (n) = \{B \in \mathbb{R}^{n \times n} | B^TB = I, det (B) = 1\},
    \end{equation}
and the unitary group formulated as
    \begin{equation}
    U (n) = \{B \in \mathbb{C}^{n \times n} | B^{\ast}B = I\}.
    \end{equation}
The tangent space at the identity element of the Lie group is called the $Lie \ algebra$ of it. For the special orthogonal group and the unitary group, their Lie algebras are given by
    \begin{equation}
    \begin{aligned}
    \mathfrak{so} (n) &= \{A \in \mathbb{R}^{n \times n} | A\ + A^T = 0\}, \\
    \mathfrak{u} (n) &= \{A \in \mathbb{C}^{n \times n} | A\ + A^{\ast} = 0\}.
    \end{aligned}
    \end{equation}
$\mathfrak{so} (n)$ is known as skew-symmetric matrix, while $\mathfrak{u} (n)$ is skew-Hermitian matrix.
The Lie exponential map  ($exp: \mathfrak{g} \rightarrow G$ where $G$ denotes the Lie Group and $\mathfrak{g}$ denotes its  Lie algebra) on a connected, compact Lie group is surjective. Therefore, the optimization problem on a Lie group can be converted to the optimization problem in  Euclidean space where Euclidean gradient descent optimizers can be directly used. 
    

\section{Applications in Classical Machine Learning}
\label{sec:03}
Classical machine learning methods gained achievements  in solving artificial intelligence problems (e.g., dimension reduction, inverse problem, sparse representation, analysis operator learning, and temporal models).
Despite the increasing computing power of modern computer facilities, it is still difficult to solve a large category of constrained classical machine learning problems in Euclidean space.
To decrease the solving difficulty, geometric optimization focuses on the special structure of  constrained problems and regards them as unconstrained ones on Riemannian manifolds \cite{hu2020brief}.
%
\subsection{Dimension Reduction}
By using a mapping $\mu \colon \mathbb{R}^{m} \to \mathbb{R}^{l}$ with $l < m$, dimension reduction (DR) aims to find a lower-dimensional representation $y_{i} \in \mathbb{R}^{l}$ of given data samples $x_{i} \in \mathbb{R}^{m}$.
%
The most popular DR paradigm uses a linear projection while others employ a nonlinear transformation to constrain
locality properties between data. 
Table~\ref{Tab:DR} summarizes main properties of mainstream DR  approaches (e.g., linear discriminant analysis  (LDA)  \cite{ye2004lda}, principal component analysis  (PCA)  \cite{jeffers1967two,woldprincipal,zou2018selective}, 
multi-dimensional scaling (MDS)  \cite{rehm2005mds,buja2008data}, isometric feature mapping  (ISOMAP)  \cite{fan2012isometric}, local linear embedding  (LLE)  \cite{ni2020fast},
laplace eigenmaps  (LE)  \cite{li2019survey}, and locality preserving projections  (LPP)  \cite{wang2017fast}).  

The mapping $\mu \colon \mathbb{R}^{m} \to \mathbb{R}^{l}$ used in DR methods is often restricted to be an orthogonal projection, i.e., 
\begin{equation}
    \mu (\x) := \U^{\top} \x,
\end{equation} 
where the orthogonal matrix $\U \in \mathbb{R}^{m\times l}$ belongs to the \emph{Stiefel} manifold
${St} (l,m) := \big\{ \U \in \mathbb{R}^{m\times l}| \U^{\top} \U = \I_{l} \big\}$.
%
%
One generic algorithmic framework to find an optimal $\U \in St (l,m)$
can be formulated as a maximization problem, i.e.,
\begin{equation}
\label{eq:trace_quotient_st}
	\operatorname*{argmax}_{\U \in St (l,m)} \,
	\frac{\operatorname{tr} (\U^{\top} \A \U)}{\operatorname{tr} (\U^{\top} \B \U) + \sigma},
\end{equation}
where matrices $A, B\in \mathbb{R}^{m \times m}$ are often symmetric or positive definite matrices. 
Equation~\eqref{eq:trace_quotient_st} is called \emph{trace quotient} 
or \emph{trace ratio}.
Note that constant $\sigma > 0$ can prevent the denominator from being zero.
%
%
Matrices $A$ and $B$ are constructed to measure the similarity between data points according to specific problems.
$V$ is not unique and closely related to selected eigenvalues.
%
%
%
Solutions of Equation~\eqref{eq:trace_quotient_st} are 
rotation invariant, i.e., let $\U^{*} \in St (l,m)$ be a solution of the problem, then 
$\U^{*}\Thetab$ for any orthogonal $\Thetab \in \mathbb{R}^{l \times l}$ is also a solution of Equation~\eqref{eq:trace_quotient_st}.
In other words, the solution set of Equation~\eqref{eq:trace_quotient_st} is the set of all $l$-dimensional linear subspaces in $\mathbb{R}^{m}$, which can be represented by \textit{Gra{\ss}mann} manifold
, i.e.,
\begin{equation}
\label{eq:grassmann}
	\mathfrak{Gr} (l,m) := \left\{ \U \U^{\top} | \U \in St (l,m) \right\}.
\end{equation}
As shown above, most linear DR methods begin with solving $tr (V^TAV)$  
while nonlinear DR methods construct a graph by connecting nearby points, which captures information on the local neighborhood structure of data and forms a similar optimization problem.
Taking the non-linear DR method LE as an example, the Laplace matrix associated with the neighborhood graph  \cite{kokiopoulou2011trace} can be regarded as the symmetric matrix $A$ in Equation~\eqref{eq:trace_quotient_st}. 
\begin{table*}[ht]
\begin{center}
{
\begin{threeparttable}[b]
\caption{Summary of Dimension Reduction Algorithms}
\label{Tab:DR}
\scriptsize
\begin{tabular}{|c|c|c|l|}
\hline
\rowcolor{gray!40}
{\textbf{Methods}} &
{\textbf{$Linear/Non-Linear^1$}} & {\textbf{$Global/Local^2$}}& {\textbf{Properties}} \\ \hline
\hline
LDA 
& Linear & Global & \makecell[l]{is supervised, uses prior knowledge of categories,\\is limited to Gaussian distribution samples} \\ 
\hline
PCA 
& Linear & Global & \makecell[l]{is unsupervised, uses orthogonal principal components\\ to eliminate interactions between each components} \\
\hline
MDS 
& Non-Linear & Global & \makecell[l]{has simple calculation,
preserves the data\ relationship\\ in\ original space,  is visualization-friendly, mistakenly\\ assumes that each\ dimension\ has a same contribution} \\ \hline
ISOMAP 
& Non-Linear & Global & \makecell[l]{suits\ low\ dimensional\ manifolds with\ a flat\ interior\\ rather than that\ with\ large\ internal curvature, has high\\ computation cost} \\ \hline
LLE 
& Non-Linear & Local & \makecell[l]{suits non-closed locally\ linear low\ dimensional\\ manifolds, has small computational complexity,\\ is limited to dense uniform dataset, is sensitive\ to the\\ number of nearest\ neighbor samples} \\ \hline
LE 
& Non-Linear & Local & \makecell[l]{preserves local features, is less sensitive to outliers and\\ noise, has a stable embedding}\\ 
\hline
LPP 
& Linear & Local & \makecell[l]{is defined\ at\ any\ point\ in\ space, i.e., can be generalized\\ to the testing set and not\ limited\ to\ the\ training\ set} \\ \hline
\end{tabular}
\begin{tablenotes}
    \item[1] \textit{Linear} represents linear projection mapping, while \textit{non-linear} represents non-linear projection mapping.
    \item[2] The \textit{global/local} represents the geometric relationship of the input data. 
\end{tablenotes}
\end{threeparttable}}
\end{center}
\end{table*}
\subsection{Inverse Problem}
Aiming to explore internal patterns from phenomena  \cite{tarantola2005inverse}, an inverse problem has a significant impact on practical applications. 
For example, the following practical problems can be modeled as inverse problems: i) deducing structural information in human body from the X-ray; and ii) infering interior appearance of stratigraphy from seismic wave.
An inverse problem can be viewed as reconstructing inputs from outputs as follows,  
\begin{equation}  
    \label{eq_2_2_inverse_problem}
   \mathbf{y} = \bm{{W}} {\mathbf{x}} 
   ,
\end{equation}
where 
$\mathbf{y}\in \mathbb{R}^l$ is the given output and $\bm{{W}}$ is a matrix that maps input data $\mathbf{x}$ to output data $\mathbf{y}$.
The goal of the inverse problem in  Equation~\eqref{eq_2_2_inverse_problem} is to recover $\mathbf{x}$ on the premise that $\mathbf{y}$ is a priori.
It is challenging to get a precise solution, however, an approximate solution 
can be achieved by confining the parameter matrix $\bm{{W}}$ to reside on a smooth Riemannian manifold. 
Let the sum of elements in the same row of matrix $\bm{{W}}$ be exact 1, Equation~\eqref{eq_2_2_inverse_problem} can be solved by  optimization  on the oblique manifold $\mathcal{M}$ where  matrices all have unit row sums  \cite{2021Variational}, i.e.,
\begin{equation}
    \min_{\bm{{W}} \in \mathcal{M}} \left \|\mathbf{y}-\bm{{W}} \mathbf{x} \right \|_{2}^{2}.
\end{equation}
\subsection{Dictionary Learning}   
As a specific inverse problem, dictionary learning has been widely used to obtain the most essential features of input data \cite{hawe2013separable_cvpr}. 
Let $X\in R_{n\times k}$ denote the input sample, in dictionary learning, $X$ is expanded into a linear combination as
\begin{equation}  
\label{eq_3_1}
    X = D_{1}\phi_{1} + \cdots + D_{n}\phi_{n},
\end{equation}
where $D_{1},\cdots,D_{n}$ represent the most essential features to be learned from the input, while $\phi_{1},\cdots, \phi_{n}$ indicate combination coefficients of features $D_{1},\cdots,D_{n}$.
Let $D\in R_{k\times n}$ indicate the dictionary set $\{D_{1},\cdots,D_{n}\}$ and $\Phi  \in R_{n\times r}$ indicate the set $\{\phi_{1},\cdots, \phi_{n}\}$, Equation~\eqref{eq_3_1} can be simplified as follows,
\begin{equation}
    X = D\Phi,
\end{equation}
where $D$ and $\Phi$ can have various kinds of combinations. 
Dictionary learning aims to learn a $D$ that makes the coefficients $\Phi$ be zero or close to zero, i.e., a sparse representation of samples $X$.
    %
The dictionary $D$ and the sparse coefficients $\Phi$ are calculated alternately.
When $\Phi$ is fixed, the dictionary learning part is the same as the form of Equation~\eqref{eq_2_2_inverse_problem}, which is an inverse problem of reconstructing $D$.
    Let $\left \|\phi \right \|_{0}$ denote the number of entries in $\Phi$ that are different from zero, the dictionary $D$ is subject to $\left \|D_{1} \right \|=...=\left \|D_{n} \right \|=1$.
    Therefore, the above dictionary learning problem can be transformed to the following minimization problem on the oblique manifold:
    \begin{equation}  
    \label{eq_3_2}
    	\argmin_{D\in \textit{OB} (k, n)} \left \|X-D\Phi \right \|_{2}^{2}+\lambda \left \|\Phi \right \|_{0}.
    \end{equation}
    
\subsection{Analysis Operator Learning}

Analysis operator learning assumes that a few operators are sufficient to represent observed high-dimensional variables  \cite{hawe2013analysis}. 
However, these operators are implicit and unobserved, for instance, store environment and service quality are latent operators hidden behind the observed variable ``price''. 
The goal of analysis operator learning is to find out these invisible operators, 
since low-dimensional operators  can simplify original high-dimensional variables. 

Let $X$ be original high-dimensional variables and $F$ be latent operators with lower dimensions, the analysis operator learning can be generally formulated as follows
\begin{equation}
    X = A F,
\end{equation}
where $A$ denotes the operator loading matrix, in which the element $A_{ij}$ represents the load of variable $x_i$ on factor $f_j$. It is proved that the parameter $A$ can be positive \cite{2004Positive}, the analysis operator learning can therefore be converted to an optimization problem on the positive manifold $\mathcal{M}$ as follows
\begin{equation}
    \min_{A \in \mathcal{M}} \left \|X-A F \right \|_{2}^{2}.
\end{equation}

\subsection{Temporal Model}
The temporal probability model is composed of a transition model describing the state evolution over time and a sensor model describing the observation process  \cite{Wei2017Dynamic_TIP}. A temporal model is helpful to cope with filtering, prediction and smoothing.
In the transition model, next state $z_{t+1}$ is transited from the current state $z_{t}$, independent from previous states. Given the time-relevant transition probability $A (t)$, the transition process of states can be modeled as
\begin{equation}
\label{temporal_1}
    z_{t+1} = A (t) \cdot z_{t} + \epsilon (t),
\end{equation}
where the noise $\epsilon (t)$ follows the Gaussian distribution. 

States are invisible and a hidden state can manifest as a specific observation with the help of an emission probability.
The current observation $x_t$ is only defined by the current state $z_{t}$, having nothing to do with previous states and observations. Given a time-varying emission probability $C (t)$, the observation process can be modeled as
\begin{equation}
\label{temporal_2}
     x_t = C (t) \cdot z_{t} + \delta (t),
\end{equation}
where the noise $\delta (t)$ follows the Gaussian distribution. 

As a mixture of Equation~\eqref{temporal_1} and Equation~\eqref{temporal_2}, temporal models can be divided into hidden Markov models and linear dynamic systems.
A hidden Markov model has discrete hidden state variables while the hidden state and observed variables of a linear dynamic system obey Gaussian distribution. 
Let $n$ represent the size of the temporal sequence, the expectation of observation sequences $E[x_0,x_1,x_2 \cdots]$ can be deduced as:
\begin{equation}
    [C (t),C (t)A (t),C (t)A (t)^2 \cdots C (t)A (t)^{n-1}]\,z_0,
\end{equation}
where $z_0$ is the initial hidden state. It can be considered as a sequence of subspaces spanned by the emission and transition matrix columns at the corresponding time  \cite{2009temporal}. As is mentioned in Section~\ref{sec:intro_manifolds}, a point on the \textit{Gra{\ss}mann} manifold is a subspace. 
Therefore, the temporal model can be mathematically optimized on the \textit{Gra{\ss}mann} manifold.

\section{Applications in Deep Learning}
\label{sec:04}
With the increasing attention to geometric optimization, more and more deep learning methods have developed to
combine with it.
Geometric optimization techniques vary with different deep learning backbones (e.g., CNN, RNN and GNN). Therefore, 
this section classifies applications in deep learning into the following categories, i.e., i) geometric CNN; ii) geometric RNN; iii) geometric GNN and iv)  geometric optimization for other deep learning methods, such as transfer learning and optimal transport.
%
%
Orthogonal manifold is widely employed in geometric CNNs to reduce feature redundancy. 
Examples include utilizing kernel orthogonality in Orthogonal CNNs  \cite{Wang2020OrthogonalCN},
optimization on Submanifolds of Convolution Kernels in CNNs  \cite{Ozay2016OptimizationOS}, 
and regularizing the convolution kernel with orthogonality when training deep CNNs  \cite{Bansal2018CanWG}.
%
%
In addition, geometric CNNs can leverage the unique structure of \textit{Stiefel} manifold  \cite{Huang2017ARN} and \textit{Gra{\ss}mann} manifold  \cite{Huang2018BuildingDN}. 
Geometric RNNs  take advantage of the norm-keeping property of orthogonal and unitary manifolds to alleviate gradient explosion and vanishing problems. 
Examples include complex unitary matrices in Unitary Evolution Recurrent Neural Networks  \cite{2015Unitary}, and the special orthogonal group and unitary group in Cheap Orthogonal Constraints:
A Simple Parameterization of the Orthogonal and Unitary Group  \cite{lezcano2019cheap}. 
Geometric GNNs pay much attention to hyperbolic manifold and extensively use it for structure capturing. 
Examples include the hyperbolic GNN  \cite{Liu2019HyperbolicGN} and a geometric neural network which incorporates Euclidean space with hyperbolic geometry  \cite{Zhu2020GraphGI}. 
%
%

\subsection{Geometric CNN}

Deep CNN has achieved great success in various computer vision tasks, such as image recognition  \cite{2018Convolutional} and segmentation tasks \cite{2018Automatically}.
 CNN can automatically learn features from large-scale data by benefiting from three essential structures, i.e., convolution, activation, and pooling structures  \cite{hu2020brief}. 
Although CNNs have worked efficiently, using the entire Euclidean space to search optimal solutions cause problems (e.g., training instability and feature redundancy) that hinder the further development. 
To alleviate these problems, geometric optimization approaches optimize CNNs on the suitable Riemannian manifold via kernel space, geometric regularization, and quasi-CNN architectures with parameters on the manifold.
%
%
%
%

\textbf{Kernel Space.}
A low-dimensional manifold is often embedded in the high-dimensional Euclidean space. 
Kernel functions can map original features to a higher dimensional space.
Therefore, with the help of kernel functions, computationally cheap operations on manifolds can represent complex operations in Euclidean space.
Kernel spaces can be utilized and described by topological smooth manifolds.
For example, 
positive-definite kernels, which are known as \textit{Gra{\ss}mann} kernels on the \textit{Gra{\ss}mann} manifold, can be used to map the manifold into a Hilbert space  \cite{hamm2008grassmann}. 
Zhang et al.  \cite{8444452} designed a new kind of \textit{Gra{\ss}mann} kernel based on canonical correlations to  distinguish one class from others more accurately. 
Liu et al.  \cite{2014Combining} designed RBF kernels for linear subspace, covariance matrix, and Gaussian distribution to optimize emotion video recognition on the Riemannian manifolds.
Hariri et al.  \cite{hariri2020efficient} defined a kernel based on the SPD covariance matrix to indicate the similarity of two face images for face matching.

Kernel space constructed on nonlinear data helps learn the inherent manifold structure.
Yuan et al. \cite{2015SceManReg} combined manifold kernel space with deep learning architecture for scene recognition. 
To preserve the geometric structure of input scene images and achieve a greater representational ability, \cite{2015SceManReg} defines a low-level feature layer $X$ and a hidden manifold kernel space $Y$ as a base unit.
Moreover, the deep architecture is unit-by-unit and $Y_k$ serves as the input of another base unit to generate the next hidden space $Y_{k+1}$. 
Comparative experiments evaluate the performance of the manifold regularized deep network on the large-scale scene data set.

Ozay et al. \cite{Ozay2016OptimizationOS} considered the kernel estimation problem in CNNs as an optimization on embedded or immersed submanifolds of kernels. \cite{Ozay2016OptimizationOS} explores geometric properties of convolution kernel space in CNNs and reveals that different kernel normalization methods induce different geometric properties.
For example, the orthonormal normalization manner implies \textit{Stiefel} manifold, while kernels normalized with the unit-norm reside on the sphere manifold. 
Furthermore,  \cite{Ozay2016OptimizationOS} proposes an SGD algorithm for optimization on kernel submanifolds.
Experiments carried out on three kernel submanifolds confirm that the above approach can boost the performance of traditional CNN training.
%

\textbf{Geometric Regularization.}
Regularization acts as the penalty term of the optimization function.  It is used to impose restrictions on the parameters of the optimization function.
The commonly used geometric regularization is the orthogonal constraint, aiming to restrict parameters to be on the orthogonal manifold.
Recall orthogonal matrices $W^{T}W = I$ introduced in Section~\ref{sec:intro_manifolds}, orthogonal regularization methods are roughly divided into hard orthogonality as
\begin{equation}
    \Vert W^{T}W - I
    \Vert_F^2
\end{equation}
and soft orthogonality as \begin{equation}
    \lambda \Vert W^{T}W - I\Vert_F^2
\end{equation}
where $\Vert \cdot \Vert_F$ indicates the Frobenius norm and $\lambda$ represents a relaxation coefficient.
Based on the soft orthogonality, we can achieve double soft orthogonality as 
\begin{equation}
    \lambda( \Vert W^{T}W - I\Vert_F^2 + \Vert WW^{T} - I\Vert_F^2).
\end{equation}

Based on the observation that the kernel orthogonality is necessary but insufficient for the orthogonal convolution, Wang et al. \cite{Wang2020OrthogonalCN} proposed an approach where orthogonality constraints directly regularize a convolution layer.
During training, the convolution filter $K$ is transformed into a Doubly Block-Toeplitz (DBT) matrix and the spectrum is regularized to be uniform, which requires row or column orthogonality. 
The orthogonality constraint on the DBT matrix helps relieve exploding and vanishing gradient problems, making the training more stable. Moreover, a number of experiments show that it can achieve amazing performance such as stronger robustness and better generalization.

Bansal et al. \cite{Bansal2018CanWG} observed that orthogonality can stabilize the energy distribution of activations within CNNs and enhance the efficiency of training. \cite{Bansal2018CanWG} compares different orthogonality regularizers, e.g. soft orthogonality, double soft orthogonality and mutual coherence regularization that lowers the column correlation as much as possible to enforce orthogonality.
Meanwhile, \cite{Bansal2018CanWG} designs a novel orthogonality regularizer named Spectral Restricted Isometry Property Regularization, which focuses on minimizing the spectral norm of $W^{T}W - I$.
Remarkable experimental results suggest that regarding  orthogonality regularizations as standard tools for training deep CNNs offers better
accuracy and stablity.
 

In order to estimate human face poses under challenging circumtances such as complex background or various orientations, Hong et al.  \cite{hong2018multimodal} proposed manifold regularized convolutional layers (MRCL) to enhance the nonlinear locality constraints of CNN parameters. 
With MRCL being on top of traditional CNN's pooling and activation operations,
a low-rank manifold structure of latent data can be recovered for better optimization.
By employing multitask learning with low-rank learning, multimodal of different data representations can be combined to predicate face postures. Comparative experiments validate the benefit of imposing manifold regularization to traditional convolutional layers.

Roufosse et al.  \cite{roufosse2019unsupervised} proposed a spectral unsupervised functional map network  (SURFMNet) 
where the matching network from one shape to another is constrained to the orthogonal manifold.
SURFMNet computes correspondences across 3D shapes using unsupervised learning, i.e., building shape correspondences without ground truth.
Solid experimental results support the consistent superiority of SURFMNet compared to state-of-the-art unsupervised shape matching methods.
Experimental results also show that SURFMNet is comparable to supervised ones.

Different from existing methods that shallowly learn Lie group features, Huang et al. \cite{huang2017deep} incorporated a Lie group structure to parameter matrices in the deep human action recognition network. The proposed skeleton-based human model $ (V,E)$ is a binary relation, where $V$ represents a set of vertexes that consists of body joints $ (v_1,\dots ,v_N)$ and
$E$ represents a set of edges that consists of body bones $ (e_1,\dots ,e_M)$. The rotation matrix is represented by the axis-angle model based on the skeleton and forms the special orthogonal group. To preserve the Lie group structure of the input rotation matrix, the above human action recognition network is optimized on the Lie group manifold and mapped to a tangent space for the final classification. 

Similar to the above action recognition network \cite{huang2017deep}, Chen et al.  \cite{chen2017deep}
put forward a deep manifold learning (DML) framework to learn manifold information and deep representations of action videos.
\cite{chen2017deep} studies that leveraging geometry information in deep learning contributes to high accuracy and efficiency for action recognition.
To extract more expressing features, the DML framework applies a manifold regularizer on the previous layer, label information and manifold parameters.
Furthermore, adapting the DML framework to restricted Boltzmann machine can relieve the overfitting problem and improve the recognition accuracy.

\textbf{Quasi-CNN Architecture.} Kernel methods and orthogonal regularization 
do not change fundamental CNN components (e.g., convolution and pooling operations). 
Another method of applying geometric optimization to deep learning is to mimic traditional CNN architecture and establish a new architecture suitable for the manifold structure.
In this article, the above architecture is named as     quasi-CNN architecture.
Convolution and activation layers are rebuilt to induce geometric optimization in the quasi-CNN architecture.
%
To achieve this goal, parameters in the quasi-CNN architecture are designed to reside on the compact \textit{Stiefel} manifold.
%
%
%
For  a more intuitive explanation, this article takes the deep SPD matrix network  (SPDNet)  \cite{Huang2017ARN} and deep \textit{Gra{\ss}mann} neural network architecture  (GrNet)  \cite{Huang2018BuildingDN} as examples.

Let $X$ be the input, and $W$ be the transformation parameter on the compact \textit{Stiefel} manifold.
First, SPDNet is designed for optimization on the SPD manifold.
Bilinear mapping  (BiMap) layer $f_{b} = WXW^{T}$ plays the role of convolution layers in traditional CNNs. 
Based on the eigenvalue decomposition 
$X$ = $U\Sigma U$, eigenvalue rectification  (ReEig) layers $f_{r} = Umax (\epsilon I,\Sigma)U^{T}$ are designed to replace nonlinear activation layers and $\epsilon$ is the activation threshold. 
SPDNet   designs the eigenvalue logarithm  (LogEig) layer to flatten the Riemannian manifold to a flat space where classical Euclidean computations can be applied.
GrNet is designed for optimization along the orthonormal manifold.
Full rank mapping  (FRMap) layers $f_{fr} = WX$ in GrNet replace the convolution layer in traditional CNNs.
Inspired by the QR decomposition $X = QR$ where $Q$ is orthonormal, GrNet designs re-orthonormalization  (ReOrth) layer $f_{fo} = XR^{-1} = Q$ to achieve an orthonormal output. 
Unlike the LogEig layer in SPDnet, GrNet 
uses inner product $XX^{T}$ to reduce the manifold to a flat Euclidean space. 
%
After pooling operations on the resulting Euclidean data, GrNet designs orthonormal mapping  (OrthMap) layer $f_{om} = U_{1:q}$ to transform the output matrix back to the orthonormal manifold, where $U_{1:q}$ denotes the first $q$ largest eigenvectors achieved by the eigenvalue decomposition.

\subsection{Geometric RNN}
\label{sec:4.2}
RNNs are designed to process sequential data since they can capture spatial and temporal dependencies between the sequential input.
Therefore, RNN can be applied in tasks such as speech recognition, text prediction, and machine translation. 
Given an input sequence $X_\tau = {x_1,x_2,\cdots, x_\tau}$  ($x_i \in \mathbb{R}^n$) with length $\tau$, a basic RNN framework is aimed to generate the output sequence $Y_\tau = {y_1,y_2,\cdots,y_\tau}$  ($y_i \in \mathbb{R}^p$). With hidden state $h$ passed recurrently into the model at each time step, output predictions $o_i \in \mathbb{R}^p$ of the RNN are computed as follows  \cite{Pascanu2014HowTC}:
\begin{equation}
\begin{aligned} 
\label{rnn_definition}
h_{i} &= \sigma (Ux_{i} + Wh_{i-1} + b), \\
o_i &= Vh_i + c,
\end{aligned} 
\end{equation}
where $U \in \mathbb{R}^{m \times n}$ is the input weight matrix, $W \in \mathbb{R}^{m \times m}$ is the recurrent weight matrix, 
$h_{i-1} \in \mathbb{R}^m$ is the previous hidden state,
$b \in \mathbb{R}^m$ is the input bias, 
$\sigma (\cdot)$ is a pointwise nonlinearity function,
$h_{i} \in \mathbb{R}^m$ is the current hidden state, $V \in \mathbb{R}^{p \times m}$ is the output weight matrix, and $c \in \mathbb{R}^p$ is the output bias. 

\subsubsection{Orthogonal RNN  (ORNN)}
Denote $\mathcal{L}$ as the objective function to be minimized, the gradient of the loss function for the hidden state is computed as:
\begin{equation}
\label{der_rnn}
\begin{aligned} 
\frac{\partial \mathcal{L}}{\partial h_i} & = \frac{\partial \mathcal{L}}{\partial h_\tau} \cdot \frac{\partial h_\tau}{\partial h_i}  = \frac{\partial \mathcal{L}}{\partial h_\tau} \cdot \prod_{j=i}^{\tau-1} \frac{\partial h_{j+1}}{\partial h_j} = \frac{\partial \mathcal{L}}{\partial h_\tau}  (\prod_{j=i}^{\tau-1}D_{j+1}W^{T}),
\end{aligned} 
\end{equation}
where $D_{j+1} \in \mathbb{R}^{m \times m}$ is a diagonal matrix, whose entries consist of the derivate of the activation function.
The pointwise non-linearity function $\sigma (\cdot)$ in Equation~\eqref{rnn_definition} is suggested to be a rectified linear unit  (ReLU) function  \cite{glorot2011deep,nair2010rectified}, whose output has a minimum value of $0$. The input
$D_{j+1}$ has at least one non-zero entry of the derivative value for all $j$.
Taking the Euclidean $l_2-norm$ to both sides of Equation~\eqref{der_rnn}, we have: 
\begin{equation}
\label{rnn_grd}
\begin{aligned} 
 \left \| \frac{\partial \mathcal{L}}{\partial h_i} \right \|_{2}  & \leqslant  (\prod_{j=i}^{\tau-1} \left \| D_{j+1}W^{T} \right \|_{2}) \left \|                         \frac{\partial \mathcal{L}}{\partial h_\tau} \right \|_{2}  =  (\prod_{j=i}^{\tau-1} \left \| W \right \|_{2}) \left \| \frac{\partial \mathcal{L}}{\partial h_\tau} \right \|_{2} \\
\end{aligned} 
\end{equation}
 If $\Vert W \Vert _2$ is greater than one, $\left \| \frac{\partial \mathcal{L}}{\partial h_i} \right \|_{2}$ grows exponentially as the increase of $\tau$. As a result, the norm of the gradient in Equation~\eqref{rnn_grd} disclosures the well-known gradient exploding problem that hinders the RNN from training  \cite{kanai2017preventing}. 
 If $\Vert W \Vert _2$ is smaller than one, $\left \| \frac{\partial \mathcal{L}}{\partial h_i} \right \|_{2}$ declines exponentially as the increase of $\tau$, which leads to gradient vanishing problems  \cite{kanai2017preventing}.
 
 A recent line of ORNNs imposes the orthogonal constraint on the hidden-to-hidden transformation of RNN. The recurrent weight transformation matrix $W$ is restricted to be on the orthogonal manifold.
 Let $A$ be an orthogonal matrix, for each vector $X$, its norm after orthogonal transformation is:
 \begin{equation}
      (AX)^{T} (AX) = X^{T}A^{T}AX = X^{T}X,
 \end{equation}
 which means that orthogonal transformations do not change the norm of the original vector. As a result, $\left \| \frac{\partial \mathcal{L}}{\partial h_i} \right \|_{2}$ can remain invariant in ORNN when the  transformation matrix $W$ in Equation~\eqref{rnn_grd} is orthogonal. Therefore, the exploding and vanishing gradient problem of RNN can be alleviated. Moreover, orthogonal constraints can be generalized to unitary constraints  in the complex domain.

\subsubsection{Recent Advances of ORNN}

uRNN  \cite{2015Unitary} constructs a large unitary matrix by simple parametric unitary matrices, i.e., the unitary hidden-to-hidden matrix $W$ is composed as follows,
\begin{equation}
\label{complex}
\begin{aligned} 
W = D_3R_2F^{-1}D_2\Pi R_1FD_1,
\end{aligned} 
\end{equation}
where $D$ is a diagonal matrix whose diagonal element $D_{j,j} = e^{iw_j}$ is defined by the imaginary unit $i$ and parameters $w_j \in \mathbb{R}$, $R = I-\ 2\frac{vv^{\ast}}{\Vert v \Vert ^2}$ is a reflection matrix with the complex vector $v \in \mathbb{C}^n$, $\Pi$ is a fixed random index permutation matrix, and $F$ and $F^{-1}$ are the Fourier and inverse Fourier transforms.
In the matrix construction strategy like Equation~\eqref{complex}, the number of parameters, memory, and computational overhead increase slowly at approximately linear speeds. 
Therefore, the training cost of large hidden layers can be reduced.
In uRNN, a variation of the nonlinear activation ReLU named modReLU has been proposed to maintain the phase of complex-valued hidden states:
\begin{equation}
\sigma_{modReLU} (z) =\left\{
\begin{aligned}
&  (\vert z \vert + b)\frac{z}{\vert z \vert},  \quad& if\ \vert z \vert + b \geq 0 \\
& 0, \quad & if\ \vert z \vert + b \le 0
\end{aligned}
\right. 
\end{equation}
where $b$ $\in \mathbb{R}$ is a bias parameter.
uRNN   defines a matrix $U$ 
to map complex-valued hidden state $h_t$ to real-valued output for prediction. The corresponding loss function is calculated as follows 
\begin{equation}
  o_t = U\left (
\begin{aligned}
Re (h_t)\\
Im (h_t)
\end{aligned}
\right)
+ b_o,   
\end{equation}
where $b_o 
$ is the output bias, $Re (h_t)$ and $Im (h_t)$ represent the real and imaginary part of $h_t$ respectively.

However, Wisdom et al. \cite{NIPS2016_d9ff90f4} noticed that the unitary parameter construction of Equation~\eqref{complex} cannot cover all $N \times N$ unitary matrices for $N > 7$, i.e., at least one $N \times N$ unitary matrix cannot be represented in the form of Equation~\eqref{complex}.
To address this problem, \cite{NIPS2016_d9ff90f4} designs a method to measure the representation capacity of the structured $N \times N$ unitary matrix.
\cite{NIPS2016_d9ff90f4} comes up with a perspective that  the unitary matrices parameterized by $P$ real-valued parameters for $P \geq N^2$ is full-capacity, which means that it can cover all $N \times N$ unitary matrices.

Unlike generating compound orthogonal matrices with simple ones,  the Lie exponential map  can   achieve orthogonal constraint on the hidden-to-hidden transformation   \cite{lezcano2019cheap}.
The connected subjective exponential mapping $exp: \mathfrak{g}$ $\rightarrow$ G on the special orthogonal group is defined as 
\begin{equation}
\begin{aligned}
exp (A) := I\ + \ A +\ \frac{1}{2}A^2 +\ \dots .
\end{aligned}
\end{equation}
Since it is a subjection, for each hidden-to-hidden transformation matrix $W$ belonging to the special orthogonal group or unitary group, there must exist a skew-symmetric  (or skew-Hermitian matrix) $A$ that satisfies $exp (A) = W$. 
Therefore, the hidden-to-hidden transformation $h_{t+1} = \sigma  ( Wh_t\ + Tx_{t+1})$ is equivalent to $h_{t+1} = \sigma (exp (A)h_t\ + Tx_{t+1})$. 
That is, the optimization on the orthogonal or unitary manifold can be transformed to the optimization in Euclidean space. 
Consequently, classic gradient descent optimizers such as Adam can be applied to minimize the loss function as well as satisfying orthogonal constraints.
As a result, the Lie exponential map can achieve both cheap computation overhead and the mitigation of gradient exploding and vanishing problems.

Another method  \cite{2016Learning}, which is   based on the Lie group, defines a basis $\{T_j\}_{j=\{1,\cdots,n^2\}}$ and coefficients $\{\lambda_j\}_{j=\{1,\cdots,n^2\}}$ to construct the element $L \in \mathfrak{u} (n)$ as follows,
\begin{equation}
    L = \Sigma _{j=1}^{n^2} \lambda_j T_j.
\end{equation}
By using exponential mapping, the element $U$ of corresponding unitary Lie group $U (n)$ can be represented as:
\begin{equation}
    U = exp (L) = exp (\Sigma _{j=1}^{n^2} \lambda_j T_j).
\end{equation}
Furthermore, Hyland et al. \cite{2016Learning} offered an argument that the above parameterization helps  generalize unitary RNN to arbitrary unitary matrices and figure out long-memory tasks.

Learning orthogonal filters in deep neural networks  (DNN) can   be formulated as an optimization problem over multiple dependent \textit{Stiefel} manifolds  (OMDSM)  \cite{huang2018orthogonal}. 
The orthogonal linear module can substitute standard linear module in DNNs to stabilize the distributions of activation and regularize networks. 
Let $W_k$ and $b_k$ be learnable weight matrix and bias, parameter $\theta$ be $\{W_k, b_k| k = 1,2,\dots K\}$, the deep neural network can be represented as $f(x,\theta):x \rightarrow \hat{y}$, where $x$ is the input feature, and $\hat{y}$ is the output prediction of DNN. The loss function is often designed as the discrepancy between label $y$ and prediction values: $\mathcal{L} (y,f(x,\theta))$. Finally, the optimization problem is formulated as 
\begin{equation}
\begin{aligned}
\theta^{\ast} = argmin_{\theta}\mathbb{E}_{ (x,y)\in D}[\mathcal{L} (y,f(x,\theta))].
\end{aligned}
\end{equation}
OMDSM trains DNN with orthogonal weight matrix $W_k$ in each layer. Thus, the optimization problem is reformulated as 
\begin{equation}
\begin{aligned}
\theta^{\ast} = argmin_{\theta}\mathbb{E}_{ (x,y)\in D}[\mathcal{L} (y,f(x,\theta))] \\
s.t. \ W_k \in \mathbb{O}_k^{n_k \times d_k},k = 1,2,\dots K,
\end{aligned}
\end{equation}
where the matrix family $\mathbb{O}_k^{n_k \times d_k} = \{W_k \in \mathbb{R}^{n_k \times d_k}| W_kW_k^T = I\} $ is composed of multiple real \textit{Stiefel} manifolds, which is an embedded sub-manifold of $\mathbb{R}^{n_k \times d_k}$. Each independent orthogonal filter $W \in \mathbb{R}^{n \times d}$ is given by the proxy parameter $V \in \mathbb{R}^{n \times d}$ as 
\begin{equation}
    W = PV,
\end{equation}
where $n$ is the number of output channels, $d$ is the number of input channels, and $P \in \mathbb{R}^{n \times n}$ is the coefficient of the linear transformation. Firstly, $V$ is centered by $V_C$:
\begin{equation}
    V_C = V - c{1_d^T},
\end{equation}
where $c = \frac{1}{d}V1_d$ and $1_d$ is the $d$-dimension vector with all ones. 
Moreover, the eigenvalues $\wedge$ and eigenvectors $D$ of the covariance matrix $V_C{V_C}^T$ are used to construct $P$:
\begin{equation}
P = D\wedge^{-1/2}D^T.
\end{equation}
Finally, $W$ is formulated as
\begin{equation}
\label{proxy_parameter}
W = D\wedge^{-1/2}D^TV_C.
\end{equation}
%

Research has   been conducted on exploring the influence of soft orthogonal constraints  \cite{vorontsov2017orthogonality}. By allowing the diagonal elements of $S$ to float around 1, the orthogonal transformation matrix $W$ is relaxed as
\begin{equation}
    W = USV^T,
\end{equation}
where $U$ and $V$ are strict orthogonal matrices.

The above methods are mainly subject to $O (n^{3})$ time complexity or dependent on complex matrices  \cite{2016Efficient}. 
It is discovered that orthogonal matrices $W \subseteq O (2n)$ with doubled hidden size, can substitute complex or unitary matrices in $\mathbb{C}^{n\times n}$.
Inspired by the above discovery, \cite{2016Efficient} proposed to utilize Householder matrices to achieve parametrization of orthogonal transition matrices. As a result, complex matrices are unneeded and time complexity is reduced, while the effect is similar to the unitary constraint.

The norm-keeping property of orthogonal matrices may make ORNN have difficulty paying little attention to extraneous information  \cite{8668730}. To relieve this problem, Jing et al.  \cite{8668730} put forward the gated orthogonal recurrent unit  (GORU) to be unconcerned with irrelevant or noise information while learning long-term dependencies. By adding the gating mechanism, experiment results demonstrate that GORU outperforms the unitary RNN on natural language processing tasks such as question answering tasks, together with long-term dependency tasks such as denoising and copying tasks.

In summary, uRNN \cite{2015Unitary} parameterizes the unitary hidden-to-hidden matrix by composing simple unitary matrices. 
However, the above parameterization cannot cover all $N \times N$ unitary matrices.
To  make up for that, full-capacity uRNN \cite{NIPS2016_d9ff90f4} is put forward. Unlike uRNN, expRNN \cite{lezcano2019cheap} exploits the exponential map to achieve orthogonal constraints more easily. 
Furthermore, OMDSM innovatively uses re-parameterization to optimize DNN over multiple dependent \textit{Stiefel} manifolds instead of one manifold \cite{huang2018orthogonal}.
Moreover, research has explored whether and how the hard orthogonal constraints on RNN can be relaxed \cite{vorontsov2017orthogonality}.
By creatively introducing the householder matrix, the considerable time complexity of parameterizing unitary matrices can be mitigated  \cite{2016Efficient}.
Last but not least, GORU \cite{8668730} designs a forget gate, so that ORNN can pay little attention to extraneous information. 

\subsection{Geometric GNN}
GNN can be used to construct a learning network based on irregular graphs. 
Each graph is represented by vertexes and edges, which describes the relationship between vertexes.
GNN encodes vertexes as feature vectors and  models edges as a relationship matrix between vertexes. 
In GNN, graph convolution is performed between the relationship matrix and the feature matrix.
Therefore, GNN can take advantage of the graph structure and update the feature information of each vertex iteratively. 
Endowing Eucludiean GNN with hyperbolic geometry can make it superior in capturing graph structure  \cite{Liu2019HyperbolicGN}. Recently, plenty of geometric GNN research has investigated how to incorporate GNN with hyperbolic manifold to benefit from a neighborhood with a highly organized structure.

To make full use of the rich geometric information in the graph, geometry interaction learning  (GIL)  \cite{Zhu2020GraphGI} incorporates Euclidean space with hyperbolic geometry by exponential and logarithmic transformations. Moreover, learnable message passing parameters are optimized on the $M\ddot{o}bius$ manifold. 
To allow each node to determinate the importance of each geometry space freely, the GIL framework employs a flexible dual-space to model both low-dimensional regular data and complex hierarchical structures. A broad spectrum of experiments show that the GIL method is adaptative to node classification and link prediction tasks.

Observing that GCN cannot cope with changes in static structure information, 
Liu et al. \cite{liu2021human} put forward a manifold regularized dynamic graph convolutional network (MRDGCN), which integrated manifold regularization into GCN to model dynamic structure information.
MRDGCN automatically updates the structure information before convengence, which makes up for GCN's inability to remain optimal in pace with the learning process. 
Considerable comparative experiments on human activity datasets and citation network datasets evaluate that MRDGCN outperforms GCN and other semi-supervised learning methods.
\subsection{Geometric Optimization for Other Deep Learning Methods}
\textbf{Robust Time Series Prediction}. Considering that noises and outliers are inevitable and important for system modeling, Feng et al.  \cite{FENG2019179} put forward a robust manifold broad learning system  (RM-BLS) for time series prediction with large-scale noisy disturbations. RM-BLS applies low-rank constraint so that features spoiled by perturbations can be abandoned by feature selection.
Furthermore, RM-BLS can also abandon features that are not satisfied to low-dimensional manifold embedding.
In addition to the low-rank manifold, \cite{FENG2019179} also considers Stiefel manifold optimization and satisfies orthogonal constraints by Cayley transformation and curvilinear search algorithm.

\textbf{Medical Reconstruction}. Geometric optimization have played an essential role in the medical field, such as magnetic resonance imaging  (MRI) for cardiac diagnosis. 
Dynamic MR can be optimized on a low-rank tensor manifold  \cite{Ke2021DeepML} to seize the powerful temporal connection between dynamic signals. Moreover, the iterative reconstruction process is flattened to a neural network for acceleration, called dubbed Manifold-Net. 
To recover free breathing and ungated cardiac MRI data, Biswas et al. \cite{biswas2019dynamic} creatively combined CNN with smoothness regularization on manifolds  (SToRM) prior.
The Laplacian matrix $L$ in SToRM $tr (X^TLX)$ is defined on the manifold to model similarities between data beyond the ambient space.
To utilize the manifold structure and patient-specific information, data denoizing based on CNN and SToRM together with conjugate gradients (CG) step take place alternatively.
Experiments confirm that combining CNN with SToRM leads to a fast and high quality reconstruction of MRI data even when the down sampling frequency is less than 8.2s of acquisition time per slice.
 
\textbf{Transfer Learning}. To maximize the utilization of finite computing resources, transfer learning aims to reuse the neural network, which is trained for task $A$, to address a similar task $B$. 
%
%
Knowledge distillation  (KD) is intended to transfer model knowledge from a well-trained model  (teacher) to a compact model  (student) with soft labels.
%
%
%
Zhang et al.  \cite{8451626} devised an end-to-end deep manifold-to-manifold transforming network  (DMT-Net) for discriminative feature learning. However, reconstructing a more discriminative SPD manifold from the original one is challenging. DMT-Net designs a local SPD convolutional layer and the non-linear SPD activation layer to deal with it. 
Huang et al.  \cite{7987001} designed a manifold-to-manifold transformation matrix $W$ and constrained the optimization to reside on the SPD manifold. Moreover, the intra-class and inter-class dissimilarity graphs are built under $W$. Hence, they can represent local geometry structures and learn the discriminative feature of SPD data.

\textbf{Optimal Transport}. Optimal transport aims to measure the distance between two probability distributions by using transport plan $\Gamma$ and cost matrix $C$ 
, i.e.,
\begin{equation}
    \min_{\Gamma \in \Pi (\mu_1, \mu_2)} trace(\Gamma^TC),
\end{equation}
where $\Pi (\mu_1, \mu_2)$ consists of joint distributions with marginals $\mu_1$ and $\mu_2$. Supposing $\mu_1$ has $m$ points and $\mu_2$ has $n$ points, the size of both $\Gamma$ and $C$ is $m \times n$.
There are works that have explored the application of geometric optimization in optimal transport problems \cite{2009A}. 
By using the Riemannian gradient descent  (RGD) algorithm, \cite{2021Coupling} explored how to convert optimal transport problems with different regularizations to the optimization problem on the coupling matrix manifold  (CMM). 
To clarify the geometry optimization process, \cite{2021Coupling}   took   classic optimal transport problems  (e.g., the entropy-regularized \cite{2018Greedy} and power-regularized optimal transport problems \cite{2016Regularized}) as an example.
Observing that the constrained set $\Pi (\mu_1, \mu_2)$
has a differentiable manifold structure, \cite{2021Manifold} and \cite{Mishra2021ManifoldOF}  solved the optimal transport problem on a generalized doubly stochastic manifold, broadening the application of manifold geometry in non-linear optimal transport problems. 
In addition to general problems, \cite{Mishra2021ManifoldOF} discusses how to adapt the above geometric optimization framework to particular ones, such as problems with sparse optimal transport map and problems of how to learn multiple transport plans simultaneously.

\textbf{Robots}. Bayesian optimization is an important technology for robots since it is effective in solving optimization problems such as controller tuning,
policy adaptation, and robot design.
Bayesian optimization is based on the Gaussian Process that relies on domain knowledge exploration. Therefore, geometry-aware Bayesian optimization emerges as a promising paradigm that can incorporate domain geometry into the optimization algorithm.
There are many commonly used kernels in Gaussian Process, among which Matérn kernel is used to study geometry-aware Gaussian process and Bayesian optimization.
Euclidean Matérn kernel is defined as follows, 
\begin{equation}
    K(x,x^{\prime}) = exp(- \frac{\Vert x - x^{\prime}\Vert^{2}}{2\sigma^{2}}),
\end{equation}
where $\sigma$ is a free parameter. Matérn kernel is a commonly used kernel function when constructing stationary Gaussian process.
Borovitskiy et al. \cite{92bf5e62} pointed out that  generalizing the Matérn kernel
to the Riemannian manifold merely by replacing Euclidean norms $\Vert x - x^{\prime}\Vert^{2}$ with geodesic distances $d_g(x - x^{\prime})$
could not produce a well-defined kernel function.
To construct the Riemannian Matérn kernel defined by stochastic partial differential equations,
Borovitskiy et al. proposed to obtain Laplace–Beltrami eigenpairs for the specific manifold and approximate the infinite sum, which forms the basis for geometry-aware Bayesian optimization on  robotics.
However, the above method suffers from two problems \cite{Jaquier2021GeometryawareBO}, i.e.,  i) the amount of computation increases exponentially with the manifold dimension; and ii) such method is inapplicable to  non-compact manifolds.
To address these problems, Jaquier et al. \cite{Jaquier2021GeometryawareBO} observed a general expression of Matérn kernels,
which is helpful to generalize them to the torus and sphere manifold. More importantly, Matérn kernels can be generalized to non-compact manifolds (e.g., SPD matrix manifold and the Hyperbolic space) by using the general expression.

\textbf{Continual learning.} Continual learning aims to remember and use the experience of previous tasks to learn new tasks, which raises requirements for the memory ability of neural networks. Chaudhry et al. \cite{chaudhry2020continual} proposed to achieve the purpose of continual learning on the low-rank orthogonal manifold. The core idea of this method is to project the gradient into disjoint low-rank orthogonal subspace by introducing task-specific projection matrix in the last second layer, which can make the gradient between different tasks orthogonal and alleviate catastrophic forgetting. The concept of gradient orthogonality was first proposed in \cite{zeng2019continual}. The essential reason for catastrophic forgetting is that learning new tasks will affect the parameters learned on the old tasks. Updating parameters in the direction orthogonal to the gradient of the old tasks can not only learn new tasks but also keep the loss of the old tasks, which alleviates catastrophic forgetting. In the deep neural network, the chain derivation process can be approximately regarded as the linear transformation of the gradient, which will destroy the orthogonality of the gradient of the earlier layers and lead to catastrophic forgetting. To ensure the orthogonality of the gradient between different tasks, \cite{chaudhry2020continual} constrains parameters on the \textit{Stiefel} manifold, making this linear transformation an orthogonal transformation.

\section{Toolbox}
\label{sec:05}

The success of the Tensorflow platform and PyTorch framework in deep learning shows that toolboxes can conveniently help build neural networks. There are  valuable toolboxes designed for quickly setting up manifolds optimization. 
Manopt  \cite{2014Manopt}, Pymanopt  \cite{2016Pymanopt}, McTorch  \cite{2018McTorch}, and Geomstats  \cite{2018geomstats} are   classic toolboxes that implement manifold geometries and optimization algorithms. Moreover, they are user-friendly and time-saving. Table~\ref{tool} compares these toolboxes from the aspect of applicable manifolds and geometry operations.

Manopt, which is built on Matlab, is a helpful tool to handle a variety of geometry constraints 
(e.g., different manifold structures introduced in ~\ref{sec:intro_manifolds}).
A Riemannian optimization in Manopt  \cite{2014Manopt} is designed as a problem including manifold structures that the search space is confined to. The cost function, or optimization object, is included in the above optimization problem as well. 
If needed, a problem structure can also cover derivatives of the objective function. 
In Manopt, solvers are functions that give a general implementation to Riemannian optimization algorithms, including steepest-descent, conjugate-gradient, and Riemannian trust-regions algorithms. Since solvers in Manopt is designed to minimize the cost function, the cost function should be multiplied by a negative one 
if it is a maximization problem.

Pymanopt  \cite{2016Pymanopt} extends Manopt to python. 
Similar to the usage of Manopt in Matlab, a Riemannian optimization in Pymanopt should be initialized with a predefined manifold and cost function. Equipped with different solvers, the optimization process and result can be diverse. Pymanopt   covers all sorts of smooth manifolds such as the oblique manifold, sphere manifold, and  \textit{Gra{\ss}mann} manifold. 
Numerable optimization algorithms are included as solvers, for instance, trust-regions, conjugate-gradient, and steepest-descent algorithms are contained by Pymanopt.

Manopt and Pymanopt are limited to shallow learning optimizations and are not applicable to deep learning optimizations.
To fill the deficiency of Manopt and Pymanopt, McTorch has been implemented by extending Pytorch  \cite{2018McTorch}, a handy framework for deep learning. 
As a result, it implements a general solution for deep learning optimizations on the manifold. Unlike Manopt and Pymanopt, Riemannian optimization in McTorch does not need to define problems, manifolds, and solvers.
Similar to Pytorch, 
Riemannian optimization in McTorch only needs to define modules and optimizers such as Adam. Network modules inherited from $torch.nn.module$ initialize layers with  manifolds and forward functions.

Geoopt  \cite{Kochurov2020GeooptRO}, which is   implemented on top of Pytorch, has a cheaper infrastructure cost than McTorch. Extended from $torch.nn.Module.parameters$, Geoopt supports tensors and parameters on the manifold.  Moreover, Geoopt provides Riemannian optimizers, for instance, $Riemannian SGD$ and $Riemannian Adam$ are available and inherited from $torch.optim.SGD$ and $torch.optim.Adam$, respectively  \cite{Kochurov2020GeooptRO}.

Another toolbox, Geomstats, is composed of two core modules, i.e., geometry and learning  \cite{2018geomstats}. The former implements Riemannian metrics, including geodesic distance. The latter implements statistics and learning algorithms inherited from Scikit-Learn classes such as \emph{K-Means} and PCA. Compared with Geomstats, other toolboxes mentioned are less modular and lack statistical learning algorithms. 
Taking clustering, one of the classic statistical learning problems, as an example, Geomstats encapsulates the class \emph{Online K-Means} with the parameter \emph{metric}. 
To perform clustering operation, users only need to initialize the Riemannian metric and call \emph{fit} function of class \emph{Online K-Means} as they do in Scikit-Learn, which is easy and convenient.
    
TheanoGeometry   \cite{Khnel2017ComputationalAI} uses Theano, a python-based and research-oriented framework, to implement differential geometry and non-linear statistics problems. 
TheanoGeometry outperforms other manifold toolboxes since it can handle symbolic calculations. Thus, Theano code can be generated from symbolic expression directly, where non-linear symbolic statistics can
be optimized with a trivial amount of code.
TheanoGeometry goes further beyond efficient symbolic computation. It implements Riemannian geometry such as geodesic equations, parallel transport, and curvature with automatic differentiation features  \cite{Khnel2019DifferentialGA}. 

\begin{table}[h]
\centering
\caption{Toolboxes Comparison in Terms of Manifolds and Geometry}\label{tool}
\begin{tabular}{|c|l|l|}
\hline 
\rowcolor{gray!40}
\multicolumn{1}{|c|}{{\bf Toolboxes}} 
& \multicolumn{1}{|c|}{{\bf Manifolds} } & \multicolumn{1}{|c|}{ {\bf Geometry}}\\
\hline 
\hline 

Manopt  \cite{2014Manopt} &\makecell[l]{Euclidean manifold,\\ symmetric matrices,\\ sphere, complex circle,\\ SO (n), \textit{Stiefel},\\ \textit{Gra{\ss}mannian}, oblique\\ manifold, SPD (n),\\fixed-rank PSD matrices}&\makecell[l]{Exponential and\\ logarithmic maps,\\tangent space\\projector, retraction,\\ vector transport,\\ egrad2rgrad,\\ ehess2rhess,\\vector, metric,\\ distance, norm}\\
\hline 
Pymanopt   \cite{2016Pymanopt} & Same as Manopt & Same as Manopt \\
\hline 
McTorch  \cite{2018McTorch} &Stiefel, SPD (n) & Same as Manopt \\
\hline 
Geoopt  \cite{Kochurov2020GeooptRO} &\makecell[l]{Euclidean manifold,\\ sphere, \textit{Stiefel},\\ Poincaré ball}&\makecell[l]{Same as Manopt}\\
\hline
Geomstats  \cite{2018geomstats} &\makecell[l]{Euclidean manifold,\\ Minkowski and\\ hyperbolic space,\\ sphere, SO (n), SE (n),\\ GL (n), \textit{Stiefel},\\ \textit{Gra{\ss}mann}ian, SPD (n),\\ discretized curves,\\ Landmarks}&\makecell[l]{Exponential and\\ logarithmic maps,\\parallel transport,\\  inner product,\\ distance, norm,\\Levi-Civita conne-\\ction, geodesics,\\ invariant metrics} \\
\hline
TheanoGeometry  \cite{Khnel2017ComputationalAI} &\makecell[l]{Sphere, ellipsoid,\\ SPD (n), Landmarks,\\ GL (n), SO (n), SE (n)}&\makecell[l]{Inner product,\\ exponential and \\logarithmic maps,\\ parallel transport,\\ Christoffel symbols,\\ Riemann, Ricci and\\ scalar curvature,\\ geodesics,\\ Fréchet mean}\\
\hline
\end{tabular}
\end{table}

\section{Performance Evaluation}
\label{sec:06}
Table~\ref{ORNN}, ~\ref{ER},~\ref{AR},~\ref{FR},~\ref{SR} 
compare the performance of aforementioned geometric optimization methods on various visual tasks (e.g., character recognition, emotion recognition, act recognition, and scene recognition tasks).
Each image dataset used in different visual tasks is summarized in Table~\ref{dataset}.
\begin{table}[h]
\centering
\caption{Datasets for Different Visual Tasks}
\label{dataset}
\begin{tabular}{|c|c|c|c|c|}
\hline 
\rowcolor{gray!40}
 {\bf Vision Task}& {\bf Dataset}&{\bf Total Samples} & {\bf Categories} & {\bf Image Size}\\
\hline 
\hline 
Character Recognition & MNIST\cite{Hdm05data} & 70000 & 10 & 32 $\times$ 32\\
\hline 
\multirow{3}*{Emotion Recognition} & 
AFEW \cite{dhall2014emotion} &1345 &7& 400 $ \times $ 400\\
& NA
BU3DFE  \cite{1613022} & 2500 & 6 & NA\\
& Bosphorus dataset  \cite{savran2008bosphorus} & 4666 & 6 &NA\\
\hline
Action Recognition & HDM05  \cite{Hdm05data} & 18000 &130 & 93 $\times $93\\
\hline
Face Verification & 
PaSC  \cite{beveridge2013challenge} & 12529
&NA
& $401 \times 401$\\
\hline
\multirow{3}*{Scene Recognition} &
Scene15  \cite{fei2005bayesian} & NA&15 & $300 \times 250$ \\
&Eight sports event categories\cite{li2007and}&NA &8 &NA \\
&
SUN  \cite{xiao2010sun, xiao2016sun}& 899&NA & NA\\
\hline
\end{tabular}
\end{table}
Table~\ref{ORNN} shows that GORU  \cite{8668730} outperforms other ORNNs on the MNIST dataset. GORU adds a forget gate, which enables ORNN to filter out irrelevant information. 
Taking advantage of the surjective exponential map, expRNN \cite{lezcano2019cheap} realizes orthogonal parameterization with a more straightforward way. Unlike expRNN, uRNN \cite{2015Unitary} uses simple unitary matrices to construct the unitary hidden-to-hidden matrix. However, such matrix construction method fails to represent all $N \times N$ unitary matrices. Therefore, Scott Wisdom et al. \cite{NIPS2016_d9ff90f4} proposed full-capacity uRNN to overcome that bottleneck of uRNN.
Using regularization terms to realize orthogonal parameterization, soRNN \cite{vorontsov2017orthogonality} explores the effect of soft orthogonal constraints on RNN.
ORNN \cite{2016Efficient} exploits the  householder matrix to enforce an orthogonal constraint on RNN, which mitigates the considerable time complexity of unitary matrices. Table~\ref{ORNN} shows that combining the forget gate, or noise filter, with ORNN improves the performance of ORNN.

\begin{table}[ht]
\centering
\caption{Comparison Results of Character Recognition}
\label{ORNN}
\begin{tabular}{|c|c|c|}
\hline 
\rowcolor{gray!40}
{\bf Dataset}& {\bf Method} & {\bf Accuracy}\\
\hline 
\hline 
\multirow{6}*{MNIST  \cite{Hdm05data}} 
& uRNN  \cite{2015Unitary} & 97.6\% \\
	~ & full-capacity uRNN  \cite{NIPS2016_d9ff90f4} & 96.9\% \\
 		~ & expRNN  \cite{lezcano2019cheap} & 98.7\% \\
 		~ & soRNN  \cite{vorontsov2017orthogonality} & 97.3\% \\
 		~ & ORNN  \cite{2016Efficient} & 97.2\% \\
		~ & GORU  \cite{8668730} & 98.9\% \\
\hline
\end{tabular}
\end{table}

\begin{table}[h]
\centering
\caption{Comparison Results of Emotion Recognition}\label{ER}
\begin{tabular}{|c|c|c|}
\hline 
\rowcolor{gray!40}
{\bf Dataset}& {\bf Method} & {\bf Accuracy}\\
\hline 
\hline 

\multirow{9}*{AFEW  \cite{dhall2014emotion}} & 
STM-ExpLet  \cite{liu2014learning} & 31.73\%  \\
		~ & RSR-SPDML  \cite{harandi2014manifold}& 30.12\% \\
		~ & DeepO2P  \cite{ionescu2015matrix} & 28.54\% \\
		~ & DCC  \cite{kim2007discriminative} & 25.78\% \\
		~ & GDA  \cite{hamm2008grassmann} & 29.11\% \\
		~ & GGDA  \cite{hamm2008grassmann} & 29.45\% \\
		~ & PML  \cite{huang2015projection_cvpr} & 28.98\% \\
		~ & SPDNet  \cite{Huang2017ARN} & 34.23\% \\
		~ & GrNet   \cite{Huang2018BuildingDN} & 34.23\% \\
\hline
\multirow{5}*{BU-3DFE  \cite{1613022}} & Tree-PNN  \cite{soyel2010optimal} & 93.23\% 
\\
		~ & Berretti et al.  \cite{berretti2010set} &  77.53\% 
		\\
		~ & Huynh et al.  \cite{huynh2016convolutional} & 92.73\%
		\\
		~ & Azazi et al.  \cite{azazi2015towards}& 85.71\% 
		\\
		~ & Hariri et al.  \cite{hariri2020efficient} & 93.50\%
		\\
		
\hline
\multirow{5}*{}
	    ~ & CSLBP  \cite{chun2013facial}& 76.98 \%
	    \\
	    ~ & CLBP  \cite{6643729}& 76.56\%
	    \\
Bosphorus & ZernikeMoments  \cite{6116669}
	    & 60.53\%
	    \\ 
	    ~  \cite{savran2008bosphorus}  
	    & Azazi et al.  \cite{azazi2015towards}
	    & 84.10\% 
	    \\
		~ & Hariri et al.  \cite{hariri2020efficient} 
		& 90.01\% 
		\\

\hline
\end{tabular}
\end{table}

Table~\ref{ER} shows that SPDNet   \cite{Huang2017ARN} and GrNet  \cite{Huang2018BuildingDN} can achieve better classification results than state-of-the-art methods on AFEW dataset  \cite{dhall2014emotion}.
The following methods for comparison are shallow learning methods applying manifold structure: Expressionlets on Spatio-Temporal Manifold  (STM-ExpLet)  \cite{liu2014learning}, 
Riemannian Sparse Representation combining with Manifold Learning on the manifold of SPD matrices  (RSR-SPDML)  \cite{harandi2014manifold}, 
Discriminative Canonical Correlations  (DCC)  \cite{kim2007discriminative}, 
\textit{Gra{\ss}mann} Discriminant Analysis  (GDA)  \cite{hamm2008grassmann}, 
Grassmannian Graph-Embedding
Discriminant Analysis  (GGDA)   \cite{hamm2009extended}, and 
Projection Metric Learning  (PML)  \cite{huang2015projection_cvpr}. 
Deep Second-order Pooling  (DeepO2P)  \cite{ionescu2015matrix} is a traditional CNN model using the standard optimization method. 
%
SPDNet exploits the \textit{Stiefel} manifold parameterization by BiMap layers and introduces non-linearity into the network by ReEig layers. 
Experiments prove that using the manifold geometry in deep learning optimization 
can improve network performance. The LogEig layer is   crucial to Riemannian computing and contributes to the emotion classification success of SPDNet. 
The success of GrNet   shows that optimizing on the \textit{Gra{\ss}mann} manifold and building a geometry-aware deep learning network is significant for
learning representative features and classifying emotions with a relatively high level of accuracy.

Table~\ref{ER}   presents that the manifold-based classification method
proposed by Hariri et al. \cite{hariri2020efficient} 
achieves the highest precision on BU-3DFE and Bosphorus datasets.
Hariri et al. used a Graph-Matching kernel and classified facial expression data with SPD covariance descriptors. It outperforms Tree-PNN  \cite{soyel2010optimal} and XP Huynh  \cite{huynh2016convolutional} on the BU-3DFE dataset by a narrow margin, and the latter two methods use traditional CNN. 
The manifold-based method proposed by Hariri et al.   greatly exceeds the methods proposed by Stefano Berretti  \cite{berretti2010set} and Amal Azazi  \cite{azazi2015towards} by approximately 15\% and 8\% on BU-3DFE dataset. In particular, the latter two methods apply SIFT and Speed Up Robust Features descriptors. 
On the Bosphorus dataset, the classification accuracy of Hariri et al.'s method \cite{hariri2020efficient} is almost far higher than all state-of-the-art methods. For example, it is even 30\% better than the  ZernikeMoments  \cite{6116669}.
%
The low-accuracy methods use local features rather than SPD covariance matrices.
Overall, these results indicate that using geometry constraints is vital for feature representation and emotion recognition.

Table~\ref{AR} shows that SPDNet achieves the highest accuracy on the action recognition task, followed by GrNet. As Table~\ref{FR} shows, SPDNet and GrNet outperform state-of-the-art methods on the face recognition task. The eigenvalue decomposition in SPDNet introduces non-linearity and the QR decomposition in GrNet performs re-orthonormalization, both of which contribute to the classification accuracy. Therefore, using matrix decomposition is vital for exploring manifold constrained parameters. 
The success of the deep manifold network on the action recognition and face recognition task shows that optimizing deep learning on the manifold helps learn favorable features and classify human actions better. 

\begin{table}[h]
\centering
\caption{Comparison Results of Action Recognition}\label{AR}
\begin{tabular}{|c|c|c|}
\hline 
\rowcolor{gray!40}
{\bf Dataset}& {\bf Method} & {\bf Accuracy}\\
\hline 
\hline 
\multirow{7}*{HDM05  \cite{Hdm05data}} 
& RSR-SPDML  \cite{harandi2014manifold} & 48.01\% \\ 
		~ & DCC  \cite{kim2007discriminative} & 41.74\% \\ 
		~ & GDA  \cite{hamm2008grassmann} & 46.25\% \\ 
		~ & GGDA  \cite{hamm2009extended} & 46.87\% \\ 
		~ & PML  \cite{huang2015projection_cvpr} & 47.25\% \\ 
		~ & SPDNet  \cite{Huang2017ARN} & 61.45\% \\ 
		~ & GrNet   \cite{Huang2018BuildingDN} & 59.23\% \\ 
\hline
\end{tabular}
\end{table}

\begin{table}[h]
\centering
\caption{Comparison Results of Face Recognition}\label{FR}
\begin{tabular}{|c|c|c|}
\hline
 \cellcolor{gray!40} & \multicolumn{2}{c|}{\cellcolor{gray!40} \textbf{Accuracy}} \\
\hhline{~|-|-|}
\rowcolor{gray!40}
\multirow{-2}{*}{ \textbf{Method}}  & \cellcolor{gray!40} PaSC1 \cite{beveridge2013challenge}
 & \cellcolor{gray!40} PaSC2  \cite{beveridge2013challenge} \\ 
\hline
\hline
VGGDeepFace  \cite{parkhi2015deep} & 78.82\% & 68.24\% \\
DeepO2P  \cite{ionescu2015matrix} & 68.76\% & 60.14\% \\
DCC  \cite{kim2007discriminative} & 75.83\% & 67.04\% \\
GDA  \cite{hamm2008grassmann} & 71.38\% &  67.49\% \\
GGDA  \cite{hamm2009extended} & 66.71\% & 68.41\% \\
PML  \cite{huang2015projection_cvpr} & 73.45\% & 68.32\%  \\
SPDNet  \cite{Huang2017ARN} & 80.12\% & 72.83\%  \\
GrNet  \cite{Huang2018BuildingDN} & 80.52\% & 72.76\% \\ \hline
\end{tabular}
\end{table}

As shown in Table~\ref{SR}, Scene Recognition by Manifold Regularized Deep Learning Architecture (SRMR)  \cite{2015SceManReg} outperforms state-of-the-art non-manifold methods on all three scene recognition datasets.  Lazebnik et al.  \cite{lazebnik2006beyond} partitioned images into fine subregions for image matching. 
Dixit et al.  \cite{dixit2011adapted} formulated Bayesian adaptation for scene image classification. 
Kwitt et al.  \cite{kwitt2012scene} recognized scene images on the statistical  (semantic) manifold.
From the perspective of information geometry, they can consider the parameter vectors as Riemannian manifolds.
Goh et al.  \cite{goh2014learning} used SIFT descriptors and represented vectorially for image recognition.
Li et al.  \cite{li2007and} interpreted the semantic components of images. Wu and Rehg  \cite{wu2009beyond} used the
Histogram Intersection Kernel  (HIK) for sports game classification. Donahue et al.  \cite{wu2009beyond} used extracted features for novel generic tasks.  SRMR's incredible success on scene recognition tasks shows that manifold regularizations are significant for improving the classification accuracy of deep learning.
\begin{table}[h]
\centering
\caption{Comparison Results of Scene Recognition}
\label{SR}
\begin{tabular}{|c|c|c|}
\hline 
\rowcolor{gray!40}
\textbf{Dataset} & \textbf{Method} & \textbf{Accuracy}\\
\hline 
\hline 
\multirow{5}*{
Scene15  \cite{fei2005bayesian}} 
& Lazebnik et al.  \cite{lazebnik2006beyond}  
& 81.2\% 
\\
~ & Dixit et al.  \cite{dixit2011adapted} 
& 82.3\% 
\\
~ & Kwitt et al.  \cite{kwitt2012scene} 
& 85.4\% 
\\
~ & Goh et al.  \cite{goh2014learning} 
& 85.4\% 
\\
~ & SRMR  \cite{2015SceManReg} 
& 86.9\% 
\\
\hline
\multirow{4}*{\makecell[c]{Eight  sports\\event\\categories\cite{li2007and}} } & Li
et al.  \cite{li2007and}  & 73.4\% 
\\
		~ & Kwitt et al.  \cite{kwitt2012scene} & 83.0\% 
		\\
	~  & Wu and Rehg  \cite{wu2009beyond} & 84.3\% 
		\\
		~ & SRMR  \cite{2015SceManReg} & 86.1\% 
		\\
\hline
\multirow{4}*{} & Xiao et al.  \cite{xiao2010sun}  & 27.2\% 
\\
		SUN  & Kwitt et al.  \cite{kwitt2012scene} & 28.9\% 
		\\
		 \cite{xiao2010sun} &Donahue et al.  \cite{donahue2014decaf} & 30.14\% 
		\\
		~ & SRMR  \cite{2015SceManReg} & 30.3\% 
		\\
\hline
\end{tabular}
\end{table}
Experimental results vary with different network architecture settings for the same manifold constrained method. 
For example, SPDNet  \cite{Huang2017ARN} has four different architecture configurations: i) SPDNet-0BiRe without using blocks of BiMap/ReEig, ii) SPDNet-1BiRe using $1$ block of BiMap/ReEig, iii) SPDNet-2BiRe using $2$ blocks of BiMap/ReEig, and iv) SPDNet-3BiRe using $3$ blocks of BiMap/ReEig.
GrNet  \cite{Huang2018BuildingDN} has three different configurations: i) GrNet-0Block without using blocks of Projection-Pooling, ii) GrNet-1Block using $1$ block of Projection-Pooling, and iii) GrNet-2Block using $2$ blocks of Projection-Pooling.
%
These methods studied how different architecture settings affected classification accuracy. 
Note that our article follows the raw settings reported from corresponding articles. On that account, this article did not present classification accuracy under different architecture configurations.
%


\section{Conclusions and Future Work}
\label{sec:conclu}
%

In this article, a survey on recent advances in applying geometric optimization to deep learning is presented.
This article reviewed progress of optimizing deep learning networks on manifolds according to the  classification of deep learning backbones (e.g., CNN, RNN, and GNN).
%
%
%
%
In particular, this article discussed the theory and toolboxes for geometric optimization.
%
Although geometric optimization brings various advantages to deep learning methods, 
it still suffers from the following challenges. 
\begin{enumerate}[-]
    \item \textbf{Dataset-Oriented Geometric Optimization. } Various  methods (e.g., uRNN \cite{2015Unitary} and Cheap Orthogonal Constraints in Neural Networks \cite{lezcano2019cheap}) utilize small image datasets such as MNIST handwritten digits to validate the effectiveness of geometric optimization. Whether geometric optimization can achieve good performance on enormous and complicated datasets such as Penn Tree Bank (PTB) needs further research. This  prompts researchers to use more challenging datasets to verify the performance of deep learning techniques after applying geometric optimization. 
    \item \textbf{Model-Oriented Geometric Optimization. } Although optimizing deep learning networks such as CNN and RNN on the Riemannian manifold has been proven successful, geometric optimization has not been applied to all deep learning methods. For example, there is a lack of research in optimizing reinforcement learning and federated learning on manifolds, which is crucial in automatic control and privacy protection. This forces researchers to further explore the potential and benefit of optimizing more deep learning networks from a geometric perspective.
    \item \textbf{Manifold-Oriented Geometric Optimization.} 
    Manifold geometry plays an important role in geometric optimization and different manifolds have different applications.
    For instance, the orthogonal manifold can be used to alleviate feature redundancy and oblique manifold  can be utilized for optimizing dictionary learning.
    However, 
    applications of certain manifolds such as centered matrix manifold remain blank in the literature. This motivates researchers to exploit and use manifold structures for geometric optimization applications as much as possible.
\end{enumerate}
This article demonstrated that geometric optimization can grasp advantage of the geometry information of search space, speed up the optimization process, and mitigate gradient explosion and vanishing problems.
However, considering unexplored deep learning methods such as reinforcement learning, together with unused manifold structures such as centered matrix manifold, it is still a huge challenge to push the boundaries of geometric optimization in deep learning.
\bibliographystyle{unsrt}
\bibliography{main}










\end{document}